%% file: main_doublecolumn.tex
\documentclass[final,5p, times]{elsarticle}

\usepackage{amssymb}
\usepackage{subcaption}

\usepackage{algorithm,mathtools}
\usepackage[noend]{algpseudocode}
\usepackage[utf8]{inputenc}
\usepackage{url}
\usepackage{booktabs}
\usepackage[normalem]{ulem}
\usepackage{bbold}

\usepackage[dvipsnames]{xcolor}

\newcommand{\add}[1]{{#1}}
\newcommand{\remove}[1]{}

\newcommand{\adda}[1]{{#1}}
\newcommand{\removea}[1]{}

\usepackage{lineno}

\makeatletter
\def\ps@pprintTitle{%
 \let\@oddhead\@empty
 \let\@evenhead\@empty
 \def\@oddfoot{\textit{\hfill\today}}%
 \let\@evenfoot\@oddfoot}
\makeatother

\journal{Remote Sensing of Environment}

\begin{document}

\begin{frontmatter}



\title{Mapping oil palm density at country scale: An active learning approach}


\author[label1]{Andr\'es C. Rodr\'iguez}
\author[label1]{Stefano D'Aronco}
\author[label1]{Konrad Schindler}
\author[label1,label2]{Jan D. Wegner}

\address[label1]{EcoVision Lab - Photogrammetry and Remote Sensing, ETH Zurich, Switzerland}
\address[label2]{Institute for Computational Science, University of Zurich, Switzerland}
\address{
\add{andres.rodriguez@geod.baug.ethz.ch}
}

\begin{abstract}
\linenumbers
\add{Accurate mapping of oil palm is important for understanding its past and future impact on the environment. We propose to map and count oil palms by estimating tree densities per pixel for large-scale analysis. This allows for fine-grained analysis, for example regarding different planting patterns. To that end,}
we propose a new, active deep learning method to estimate oil palm density at large scale from Sentinel-2 satellite images, and apply it to generate complete maps for Malaysia and Indonesia.
What makes the regression of oil palm density challenging is the need for representative reference data that covers all relevant geographical conditions across a large territory.
Specifically for density estimation, generating reference data involves counting individual trees. To keep the associated labelling effort low we propose an active learning (AL) approach that automatically chooses the most relevant samples to be labelled. 
Our method relies on estimates of the epistemic model uncertainty and of the diversity among samples, making it possible to retrieve an entire batch of relevant samples in a single iteration. %
Moreover, our algorithm has linear computational complexity and is easily parallelisable to cover large areas.
We use our method to compute the first oil palm density map with $10\,$m \add{Ground Sampling Distance} (GSD) , for all of Indonesia and Malaysia and for two different years, 2017 and 2019.
The maps have a mean absolute error of $\pm$7.3 trees/$ha$, estimated from an independent validation set.
We also analyse density variations between different states within a country and compare them to official estimates.
According to our estimates there are, in total, \textgreater1.2 billion oil palms in Indonesia covering \textgreater15 million $ha$, and \textgreater0.5 billion oil palms in Malaysia covering \textgreater6 million $ha$.
\end{abstract}





\end{frontmatter}




\input{1_introduction}
\input{2_methods}

\input{3_experiments}

\input{4_conclusion}

\section*{Credit author statement}

A.C.R was responsible for the design and development of the methods, running experiments and writing the paper. S.D., K.S. and J.D.W. were responsible for creating the research design, writing and editing. All authors discussed the results and contributed to the final manuscript

\section*{Acknowledgments}

The project received funding from Barry Callebaut Sourcing AG, as a part of a Research Project Agreement.

\clearpage

\appendix

\input{5_supplementary}





{~~~~}

\newpage

\bibliographystyle{elsarticle-num} 
\bibliography{references.bib}








\end{document}

%% file: 1_introduction.tex
\section{Introduction}
\label{sec:intro}
\add{Oil palm}
is the third largest oil crop in the world by planted area, and \add{accounted for} 35\% of the vegetable oil production in the world in 2019 \cite{usda2020oilseeds}. With the highest yield per hectare of any fat oil it is an attractive economic alternative in many tropical countries \cite{meijaard2018oil}. However, large-scale \add{oil palm} production in Malaysia and Indonesia is a \add{potential} driver of deforestation \cite{austin2019causes,gaveau2019rise}. \add{Several works relate oil palm development with} long-lasting effects on the environment, including loss of bio-diversity \cite{margono2014primary}, poor air quality and high greenhouse gas emissions \cite{noojipady2017forest,van2009co}.
On the other hand, it plays an important role for several aspects of the socio-economical life in producer countries, including positive impacts on the welfare of smallholder producers \cite{feintrenie2010farmers}.
Balancing the economic and social benefits of oil palm plantations with the impact it has on the environment is a challenging task.
We refer the reader to \cite{meijaardethicspalmoil2019} for a thorough analysis of the ethics around the palm oil industry.

Transparent, evidence-based policies for sustainable palm oil production are only possible with accurate maps at country-scale and beyond. Besides, to assess long-term impacts they shall be frequently updated. This calls for a highly automated, objective mapping process.
Several authors have studied ways to map oil palm plantations with remote sensing data, e.g., ~\cite{cheng2018towards,oon2019assessment,descals2020hiresglobalmap}.
They mostly focus on classifying the land cover into oil palms vs.\ background. Such a binary approach implicitly assumes that oil palms occur only as predominant land cover, in closed-canopy plantations with a certain minimum density, and is prone to missing smaller or less dense plantings. This may be a reason for the large variation between different estimates of planting area.
We argue that a map of oil palm density (respectively, tree count per pixel) is preferable, as it is more robust to non-uniform densities and at the same time allows for a more detailed analysis.

A general bottleneck of land cover mapping with supervised machine learning is the need for accurate, manually collected labels~\cite{marmanis2018}. Although for some applications public online data can be used, for instance from \url{openstreetmap.org}~\cite{kaiser2017}, no such data is available for more specific mapping tasks, including oil palm density. 
The standard procedure to ensure that the learned models generalise across large, geographically diverse regions is to collect a large and diverse set of manually annotated reference labels, but that brute-force approach is laborious, and therefore time-consuming and costly.

Active learning (AL) offers a possible solution, by focusing the (manual) labelling effort on a much smaller set of training examples that optimally represent the data distribution~\cite{survey2009}, thus reducing the total annotation effort required to achieve a given mapping accuracy.
The starting point for AL is the observation that naively selected training samples usually are highly redundant, such that the marginal utility of each additional sample is low.
The goal of AL is to accumulate the same evidence with a much smaller number of samples, by gradually adding new samples that are selected in an informed manner, so as to maximise their utility.
The \emph{active} construction of the training set is guided by the assumption that, in order to improve a (preliminary) prediction function, one must supply it with the correct answers for those inputs that it has not yet learned to handle. This leads to the following principles:
\begin{enumerate}
\item to decide which additional samples to label, prediction uncertainty is a useful indicator to identify samples with high marginal utility, respectively low redundancy with respect to the training data used so far.
\item if, for efficiency reasons, multiple samples are added in each round, then they should be as different from each other as possible, to avoid redundancy within the newly added batch. 
\end{enumerate}
The function that combines uncertainty and diversity into a score for selecting new samples is commonly termed the \emph{acquisition function}.

In the context of mapping, the set of unlabelled samples corresponds to the entire target region, in our case more than $\adda{10^6\,}$km$^2$ at $10\,$m ground sampling distance (GSD).
Most existing AL methods were not designed for such extremely large datasets and do not scale well enough.
For instance, many methods require that the model be retrained multiple times, and each time, it must predict the output for the entire unlabelled data set to pick new samples, which is extremely costly for large regions. Another strategy that makes it possible to extract a large and varied training set in a single shot requires the pre-computation of all pairwise similarities between candidates one may want to add in the next round~\cite{sener2017active} -- the computational complexity of that operation is obviously quadratic in the data set size.
To leverage AL for country-scale mapping, in our case of oil palms, one must design a strategy that remains viable with millions of pixels, which in practice means its computational complexity should scale (approximately) linearly with the area of the region (respectively, the number of pixels) and require as few active learning iterations as possible.

In this work we produce a country-scale map of oil palm density, by devising an active learning procedure that scales to large remote sensing datasets. The contribution of our paper is twofold:
\begin{itemize}
    \item We design a practical AL methodology for large-scale applications in remote sensing. The method relies on an implicit ensemble of randomly perturbed convolutional networks to estimate model uncertainty, together with an efficient core-set construction that uses distances between the data points' learned feature embeddings to efficiently assemble a representative and diverse training set. Our AL approach scales linearly with the number of unlabelled samples and requires only 2 processing cycles over the area of interest.
    \item By applying our AL system to Sentinel-2 imagery of Ma-laysia and Indonesia, we produce a 10$\,$m GSD, wall-to-wall map of oil palm density for those two countries, which together account for 84\% of the world's palm oil production%
    \footnote{Figure for the year 2019, according to~\cite{usda2020oilseeds}}. %
    In contrast to other \add{oil palm} mapping efforts such as~\cite{descals2020hiresglobalmap,xu2020annual,cheng2018towards} we map not only the presence/absence of oil palm plantations, but the tree density (oil palms per 100$\,$m$^\text{2}$ pixel). The map thus provides richer information for further geo-spatial analysis. For instance, it can reveal local variations in planting patterns, production intensity, and potentially relations to yield. \footnote{10m resolution maps for 2017 and 2019 are available on request.}
\end{itemize}

\section{Related work}\label{sec:related}

\begin{figure*}[!ht]
    \centering
        \includegraphics[width=\textwidth,trim={0 5.3cm 3cm 0},clip]{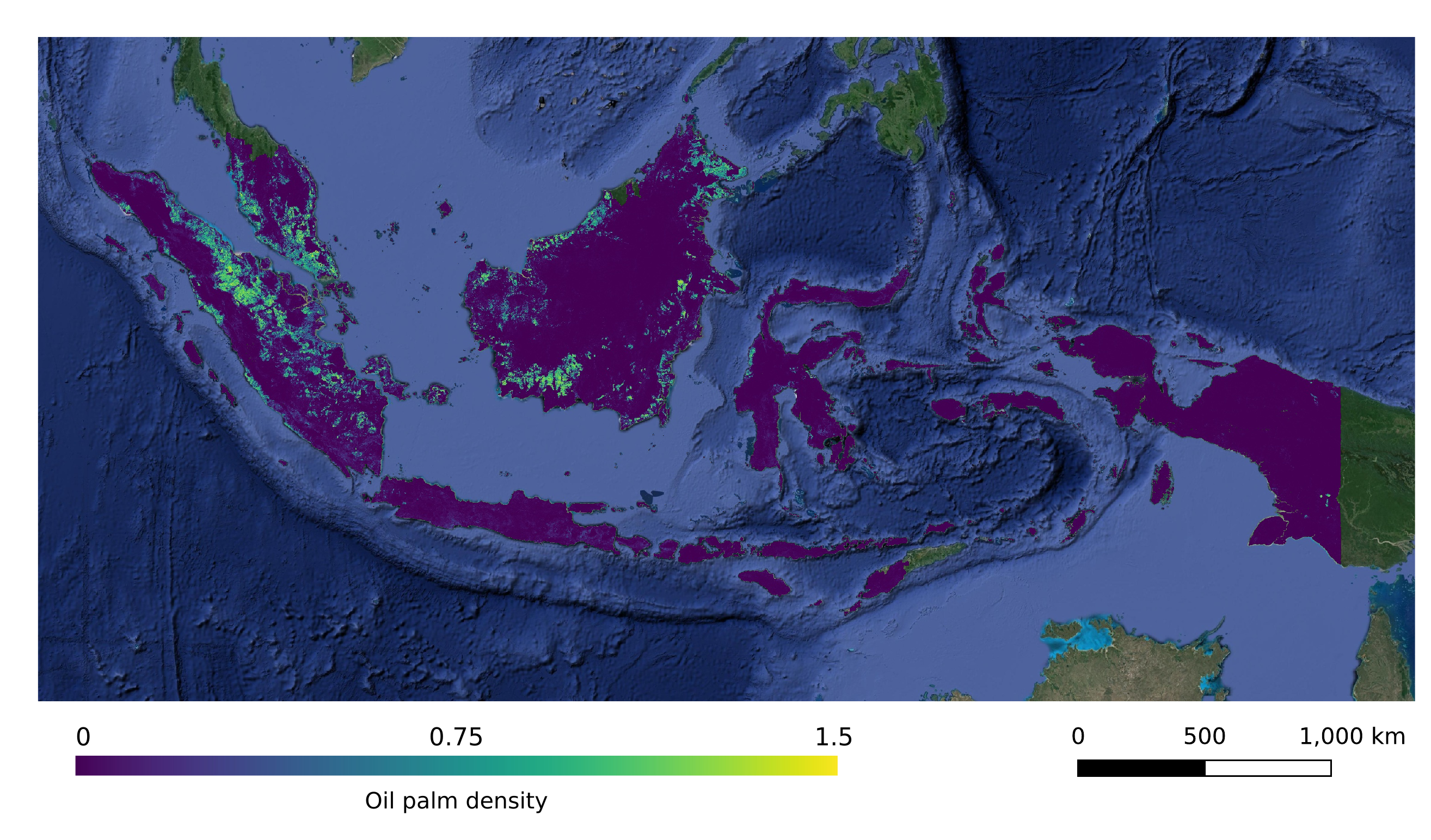}
    \caption{Oil Palm Density Map over South-east Asia in 2017 \add{using our proposed method}. \add{We estimate a total of 1.7 billion oil palms in Indonesia and Malaysia.} \adda{We present the average density over 10m pixels for illustration.}}
    \label{fig:area_study}
\end{figure*}

\subsection{Oil palm mapping}

Several authors have created maps of oil palm plantations from remote sensing data. Some use only Synthetic Aperture Radar (SAR) \cite{cheng2018mapping,oon2019assessment}, which is unaffected by the frequent cloud coverage in tropical regions \cite{cheng2018towards}.
Perhaps even more popular is the combination of optical and Radar images 
\add{
\cite{laurin2013optical,pohl2014mapping,cheng2016oil,nomura2019oil,sarzynski2020combining,descals2020hiresglobalmap,xu2020annual,gutierresannualoilpalmperu2013}.
}
Only a few works map oil palms at country-wide or continental scales. \cite{cheng2018towards} provide a oil palm map over 15 countries from around the world using, ALOS-2/PALSAR-2 data from 2016 at a 25$\,$m GSD. \cite{xu2020annual} use ALOS-2/PALSAR-2 
and MODIS to generate a map of palm oil plantations in Malaysia and Indonesia at 100$\,$m GSD, for the years 2001-2016. 
Recently, \cite{descals2020hiresglobalmap} provided a map that distinguishes between industrial and small-holder plantations. This map covers all areas suitable for palm oil production at a GSD of 10$\,$m and was derived from Sentinel-1 (SAR) and Sentinel-2 (Optical) images from 2019.

\subsection{Density estimation}

Density estimation is a recurrent task in remote sensing, including for instance the estimation of population density  \cite{robinson2017deep,doupe2016equitable} and canopy density  \cite{joshi2006remotely,rikimaru2002tropical}.
Still, \add{land cover is in most cases mapped only as a presence/absence label. Apparently species-specific} tree density estimation \add{has rarely been considered as an alternative}.
Most authors prefer to only detect trees (respectively forest or plantation areas), presumably because the size of individual trees is below the pixel footprint of wide-area satellite sensors.
However, earlier work \cite{rodriguez2018deepsemantic} has demonstrated that density estimation for sub-pixel trees is feasible.
In agricultural applications in particular, density maps allow for a more advanced analysis than binary land-cover maps of the crop. This is potentially important for oil palm plantations, where the distance between trees is known to correlate with the yield per tree \cite{bonneau2018optimum,rafii2013variation}.

\subsection{Active learning in remote sensing}

A fairly large body of literature exists about active learning methods for remote sensing. Early works often relied on Support Vector Machines (SVM)~\cite{tuia2009active,persello2014,wang2017,shi2018}. Uncertain samples are identified as those that fall within the margin of the SVM classifier (and thus also have a high likelihood to become support vectors). Those works already recognised the importance of diversity and proposed measures to avoid that multiple similar training samples \add{are} added in the same iteration of the active learning loop~\cite{tuia2009active}.
In~\cite{persello2014}, the authors analyse differences and similarities between active learning and semi-supervised learning and propose a method that combines them, leveraging the distance from a training sample to the support vectors as a measure of uncertainty to select new samples.
In \cite{sun2014} Gaussian Processes were used for classification, since they offer a natural mechanism to represent model uncertainty and, thereby, select informative samples.

Active learning with neural networks dates back to at least \cite{roy2013neural} in the remote sensing literature. While neural networks achieve excellent predictions in many image analysis tasks, they suffer from a well-known limitation that is critical in the context of AL, namely that their estimates of uncertainty are poorly calibrated. The standard work-around, already used by \cite{roy2013neural}, is to employ stochastic ensembles of models  to quantify prediction uncertainty.
To reduce the computational cost, recent active learning schemes tend to avoid explicit multi-model ensembles and instead use more efficient approximations. For instance,~\cite{haut2018active} use Monte Carlo (MC) dropout~\cite{gal2016dropout} to estimate model uncertainty. Moreover, that paper compares different acquisition functions, and confirms that significant performance gains can be achieved with active (rather than random) selection of samples.
~\cite{kellenberger2019half} combine AL with domain adaptation to detect which samples of a new \emph{target} domain need to be labelled to improve the performance of a pre-trained model on the target domain. 
They propose a method based on optimal transport that, however, relies on having a fully labelled training set in the source domain.
~\cite{Vit2020} use AL for binary change detection between co-registered image pairs. To represent model uncertainty they use Monte Carlo Batch Normalisation (MCBN)~\cite{MCBN}, another form of randomisation within the network to approximate an ensemble. Besides showing that MCBN can indeed substitute an explicit ensemble without performance penalty, the authors also show that AL tends to balance the training examples if the class distribution is highly imbalanced -- which is often the case in remote sensing when some comparatively rare target class should be separated from a dominant ``background". 
\cite{sener2017active} proposes a single shot active learning approach where a relatively large and diverse core-set of samples is extracted from the unlabelled dataset. The main characteristic of the method is that the active learning is not iterative: the core-set is extracted only once and neglects prediction uncertainty. The limitation \add{with} this approach is that the cost of building such a core-set is quadratic in the size of the unlabelled dataset.

A main challenge, without an accepted standard solution, remains the question how to scale AL to large-scale scenarios. The typical scenario in remote sensing is that the ``unlabelled data", from which one has to select the samples to be annotated, is the entire area of interest with hundreds of millions of pixels.
But most AL methods either are iterative, requiring a new prediction pass over the unlabelled set each time the model is retrained, or, \add{in order to build larger effective sets for labelling,} have a quadratic complexity with respect to the unlabelled dataset \add{(i.e., at least implicitly they form and examine all possible pairs of samples in the unlabelled set)}.
To remain tractable one would therefore have to limit them to a tiny portion of the mapping area, which, if done naively, runs the risk of missing important parts of the data distribution. We aim to solve these problems with an approximate measure of sample diversity that has computational complexity linear in the size of the unlabelled dataset, yet makes it possible to extract a diverse set of samples with few (in practice even a single) active learning iterations. 

\section{Data}

\subsection{Area of study}

This study focuses on a geographical area that comprises the countries of Malaysia and Indonesia. We produce a 10$\,$m GSD, wall-to-wall map of oil palm density for those two countries, which together account for 84\% of the world's palm oil production. This covers a land area of $1.3 \cdot 10^6 {km}^2$ or $1.3 \cdot 10^9 \text{ Sentinel-2 pixels}$. See Figure \ref{fig:area_study} for an overview of the oil palm density map. According to our estimates there are, in total, \textgreater1.2 billion oil palms in Indonesia covering \textgreater15 million $ha$, and \textgreater0.5 billion oil palms in Malaysia covering \textgreater6 million $ha$. See more details in Section \ref{sec:experiments}.

\subsection{Sentinel-2 imagery}

As input to our model, we use Sentinel-2 Level-2 bottom-of-atmosphere reflectance images acquired in 2017 and 2019. \adda{The Level-2 images were obtained using sen2cor~\cite{louis2016sentinel}.} All channels are re-sampled to a common pixel size of 10$\,$m with bicubic interpolation and stacked into 13-channel images.
First, we filter out all pixels that are labelled as cloud, cloud shadow, water or snow in the \emph{scene classification} layer provided by {sen-2cor}~\cite{louis2016sentinel}.
 Moreover, we use only pixels with $<$50\% cloud probability for training (according to the cloud mask from sen2cor), and only pixels with $<$10\% cloud probability for validation and inference. We use a different cloud probability for training to be more robust to different cloud conditions. At test time, the stricter cloud threshold yields a cleaner estimate less influenced by clouds.

To create a dense map without gaps, we process each image separately and then average all predictions at each location over the entire year.
Instead of directly mosaicing images into a single composite before processing, we process all images and perform a late fusion of results. Major advantages are that we avoid radiometric distortions from composite images with different atmospheric conditions and, more importantly, we ensure that the model learns to process images with a great variety of atmospheric conditions, such that it can be applied to any Sentinel-2 image without retraining.

We download and process Sentinel-2 images for the entire years 2017 and 2019. Using all available images allows us to compute dense, wall-to-wall maps without holes that provide an oil palm density estimate at every land pixel across Malaysia and Indonesia, despite the high cloud coverage that is common in tropical South-east Asia.

\subsection{Reference data}
\label{sec:ref_data}

To obtain reference oil palm density maps to train and evaluate our method, we rely on very high-resolution overhead imagery (GSD $<30cm$). At that resolution, individual trees are clearly visible and we thus manually annotated a total of $3.4\cdot10^6$ oil palms. All annotations were made with images acquired in 2017.
A 30-min training was required for students to learn to identify oil palm plantations and not confuse them with other crops such as coconut or sugar palms that are also present in the area of the study. Only general notions of suitable areas of oil palm plantations were required for labelling (e.g., tropical areas and elevations below 1500m \cite{pirker_limitsoilpalm2016}).
We aimed at labelling blocks of at least 250 ha each spanning large geographical areas. Manual annotators were asked to label oil palm plantations and to include as background examples other types of land cover, such as coconut plantations and forests. Inside each block, we obtained the centre location of each oil palm as observed on the very-high resolution imagery; to obtain a smooth density map at Sentinel-2 resolution (GSD 10$m$) 
we applied a smoothing operation with a squared kernel of $20\times20$ meters on a high-resolution grid (GSD 0.625 m) of the oil palm locations and then summed the densities under each Sentinel-2 pixel. This results in local oil palm density maps that can be used for training and validation of our proposed method. See more details in Section \ref{sec:large_scale_mapping} about the geographical split of train and validation areas.

%% file: 2_methods.tex
\section{Methods}\label{sec:methods}

\begin{figure}[t]
    \centering
    \includegraphics[trim={0 0 12.5cm 0},clip,width=\linewidth]{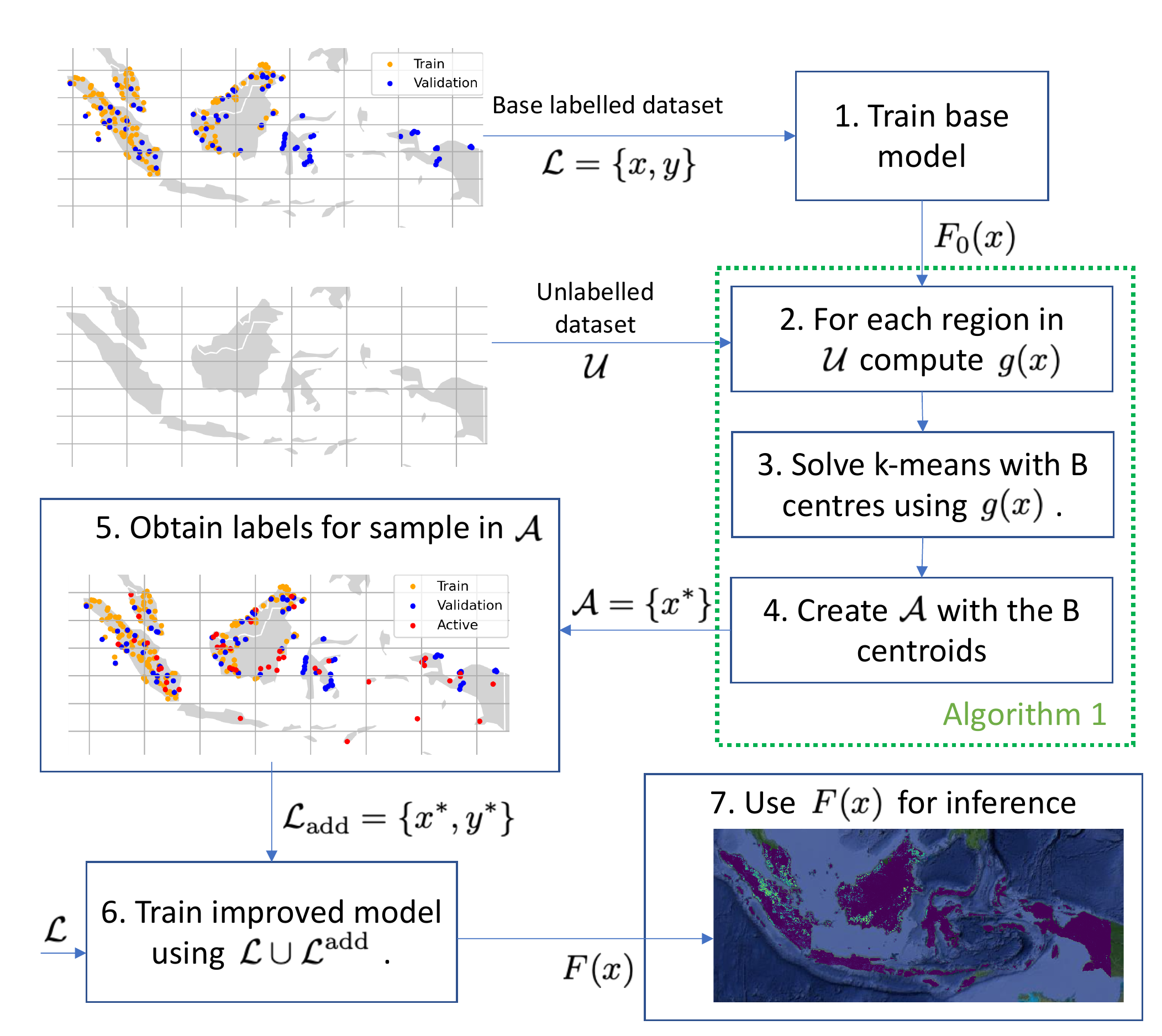}
    \caption{\add{Active Learning method overview: $F_0(x)$ is the model trained only with the base labelled dataset $\mathcal{L}$. The \emph{acquisition function} $g(x)$ can be computed in parallel for each region in the unlabelled dataset $\mathcal{U}$ and computed as described in Equation \ref{eq:sample_dist}. $g(x)$ is based on an estimate of the Epistemic Uncertainty and the distance to the centre of gravity of all data points in $\mathcal{U}$. The set $\mathcal{A}$ is the actively selected samples chosen using the Algorithm \ref{alg:core-setconstruction} in Section \ref{sec:pseudo_code}}}
    \label{fig:method_overview}
\end{figure}

\subsection{Estimation of oil palm density}

Following \cite{rodriguez2018deepsemantic}, instead of attempting to predict the locations of individual oil palms, which are smaller than the GSD of the input images, we \adda{predicted} the count of oil palms per pixel. This way, the estimation of oil palm density can be solved as a supervised regression task.
We use\adda{d} a Convolutional Neural Network (CNN), \add{$F(x)$}, that predicts the count of oil palms per pixel, $\hat{y}$, using the raw intensities of  Sentinel-2 image $x$. \add{The architecture of the CNN (i.e., layers, hyperparameters) is based on our previous work~\cite{rodriguez2018deepsemantic}.
}
\remove{Specifically, $F_\theta(x)$} \add{$F(x)$} consists of an input convolutional layer, and 15 residual layers \remove{ (each composed of 3 sequential convolutional layers).}
of the form $x_\text{out} = h(x_\text{in})+x_\text{in}$. \add{The residual layers \cite{he2016deep} learn additive updates to the input features, which has emerged as a particularly successful strategy for many computer vision and image analysis tasks. In our case, each function $h(x)$ has three sequential convolutional layers.}
\remove{Between every layer batch normalisation and a ReLU activation function are applied.}
\add{After every layer we include (\emph{i}) batch normalisation \cite{ioffe2015batch}, and (\emph{ii}) a Rectified Linear Unit (ReLU) as activation  function. Batch normalisation tracks the second-order statistics in each batch and uses them to standardise the values between different layers so as to reduce the sensitivity of the network to small changes between the batches of samples that are fed during training. ReLU is an element-wise non-linear transformation of the form $x_\text{out} = \max(0,x_\text{in})$ (it is a basic principle of neural networks to interleave linear layers with element-wise non-linearities to approximate complex functions~\cite{haykinneural}).}
Furthermore, we have two different branches as output: (\emph{i}) $y$ for the predicted count, and (\emph{i}) $y_\text{{cl}}$ for a binary classification task between pixel with oil palms and the background. 
\add{Note that at each layer we \adda{kept} the same resolution of the image from input to output, different from other architectures that reduce the resolution of the input image with an encoder and then up-sample the resolution with a decoder \cite{ronneberger2015u,chendeepLab2018,badrinarayanan2017segnet} } 
\remove{Note that we do not use any average pooling or strided operations, this}\add{. This} \adda{allowed} us to retain the input resolution and map at the original GSD without upsampling issues. See Section \ref{sec:supl_details} for more details on the architecture in the supplementary material.

To train the \add{model parameters of $F(x)$} \remove{parameters $\theta$ of $F_\theta(x)$} we use\adda{d} a reference density map that was obtained by identifying individual oil palms in high-resolution imagery \add{as described in Section \ref{sec:ref_data}}. The loss function  to train the model is a sum over the two branches,
\begin{equation}
    \text{Loss} = (y - \bar{y})^2 + CE(y_\text{cl},[\bar{y}>0])\;.
\label{eq:loss}
\end{equation}

The right-most part of Eq.~\ref{eq:loss} is the cross-entropy \add{($CE$)} between the predicted class and a binary indicator that separates background pixels with density 0 from oil palms. \add{Specifically, we \adda{used} $F(x)$ to perform a regression for each pixel in the image, where we only predict\adda{ed} the number of oil palms per pixel in the image.} For further details, refer to \add{Rodriguez et al. (2018)} \cite{rodriguez2018deepsemantic}. For simplicity, from now on we do not distinguish between the density prediction and the auxiliary classification output and simply denote the network \remove{out put} \add{output} as $y$.

\subsection{Active learning}

\def\widthfigure{0.3\linewidth}
\begin{figure*}[t!]
    \centering
    \begin{subfigure}[t]{\widthfigure}
        \centering
        \includegraphics[width=\textwidth]{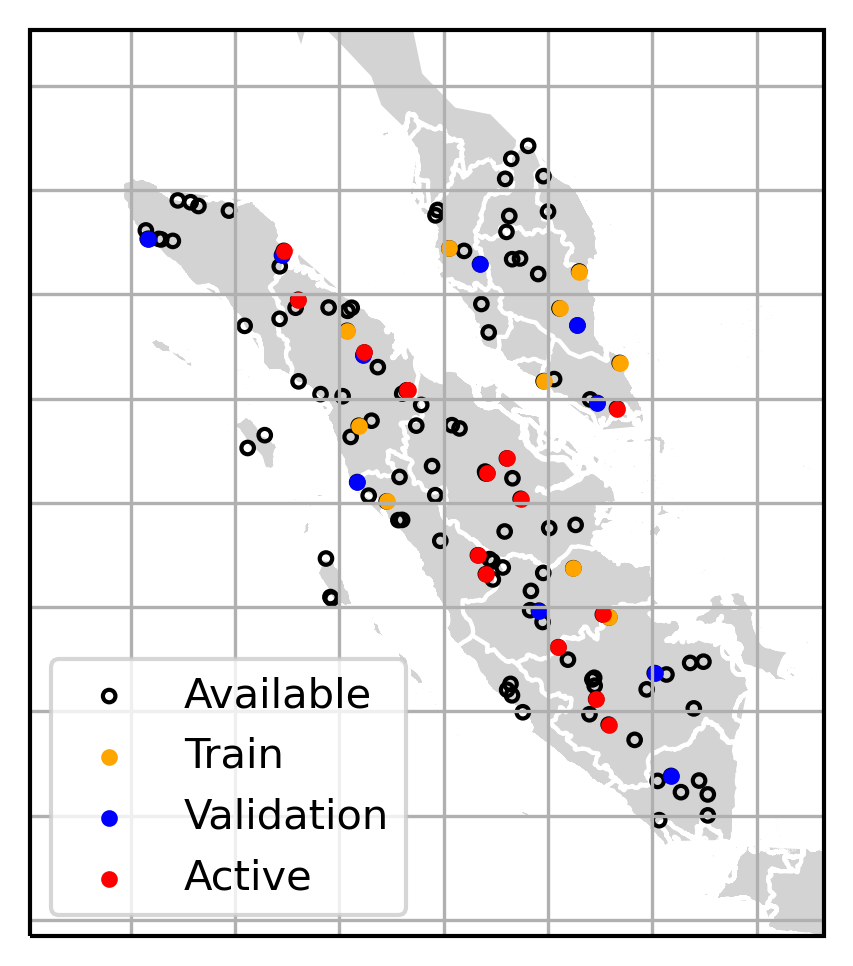}
        \caption{Active selection of 15  samples}
    \end{subfigure}
    \begin{subfigure}[t]{\widthfigure}
        \centering
        \includegraphics[width=\textwidth]{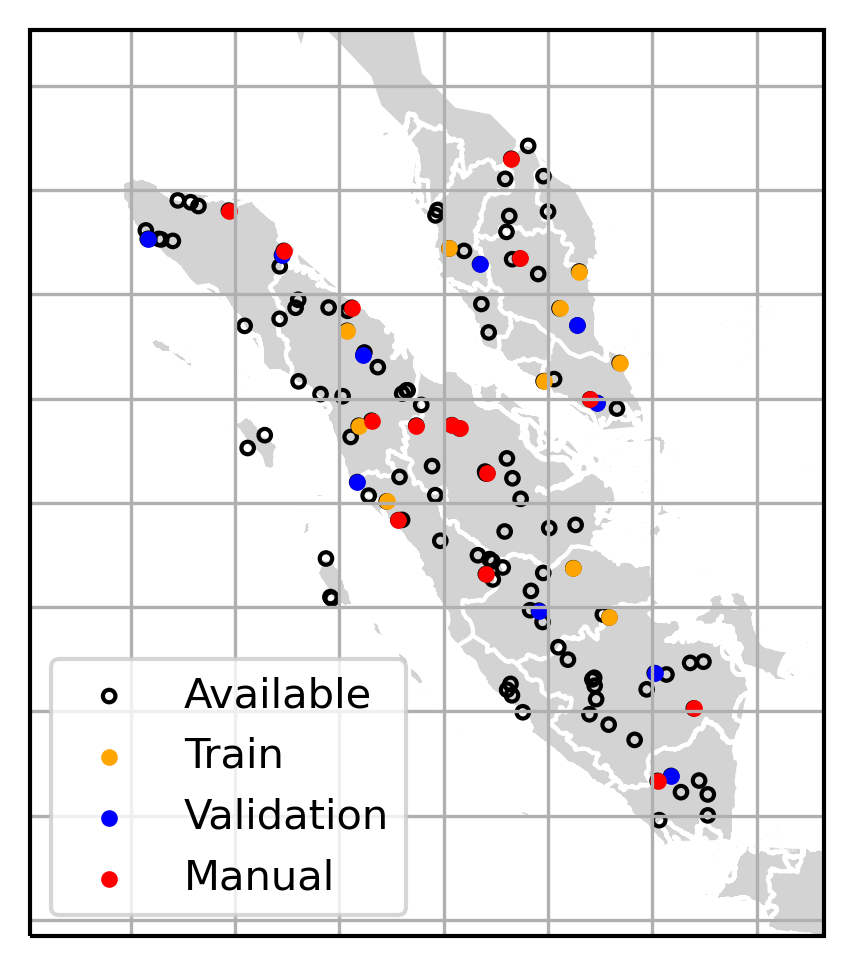}
        \caption{Manual selection of 15 samples}
    \end{subfigure}
    \begin{subfigure}[t]{\widthfigure}
        \centering
        \includegraphics[width=\textwidth]{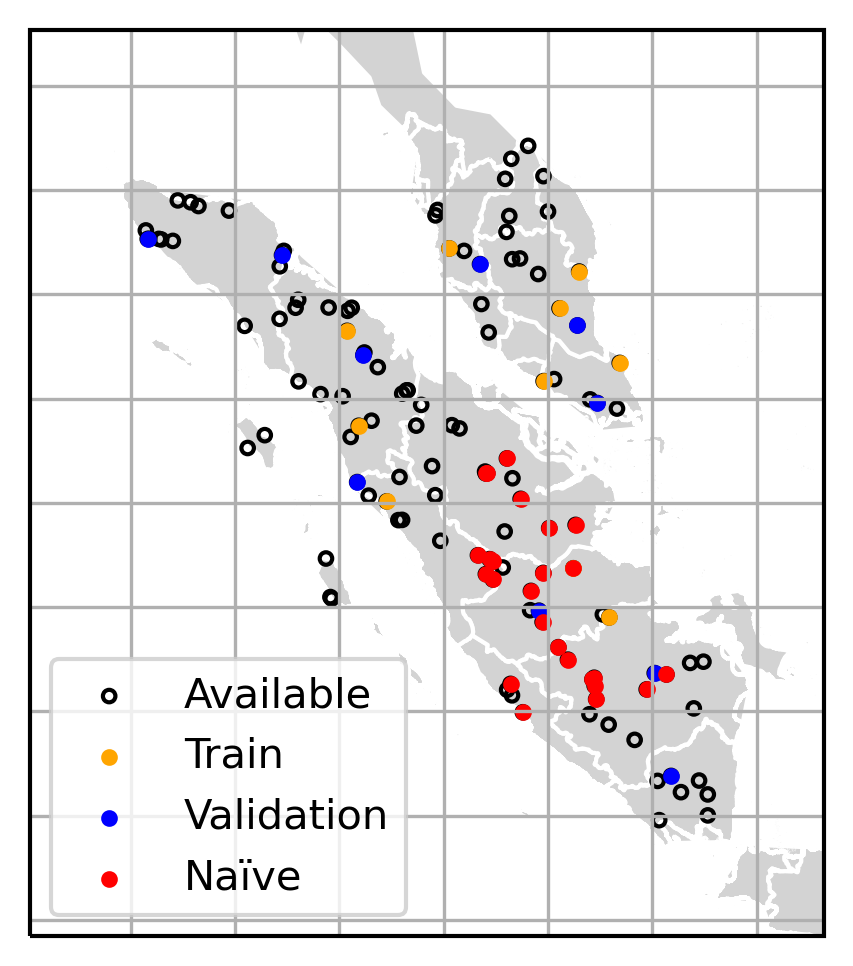}
        \caption{N\"aive selection of 15 samples}
    \end{subfigure}\quad
    \caption{Comparison of different sampling strategies from the available samples}
    \label{fig:labelled_example}
\end{figure*}

\begin{figure}[t]
    \centering
    \includegraphics[width=0.9\linewidth]{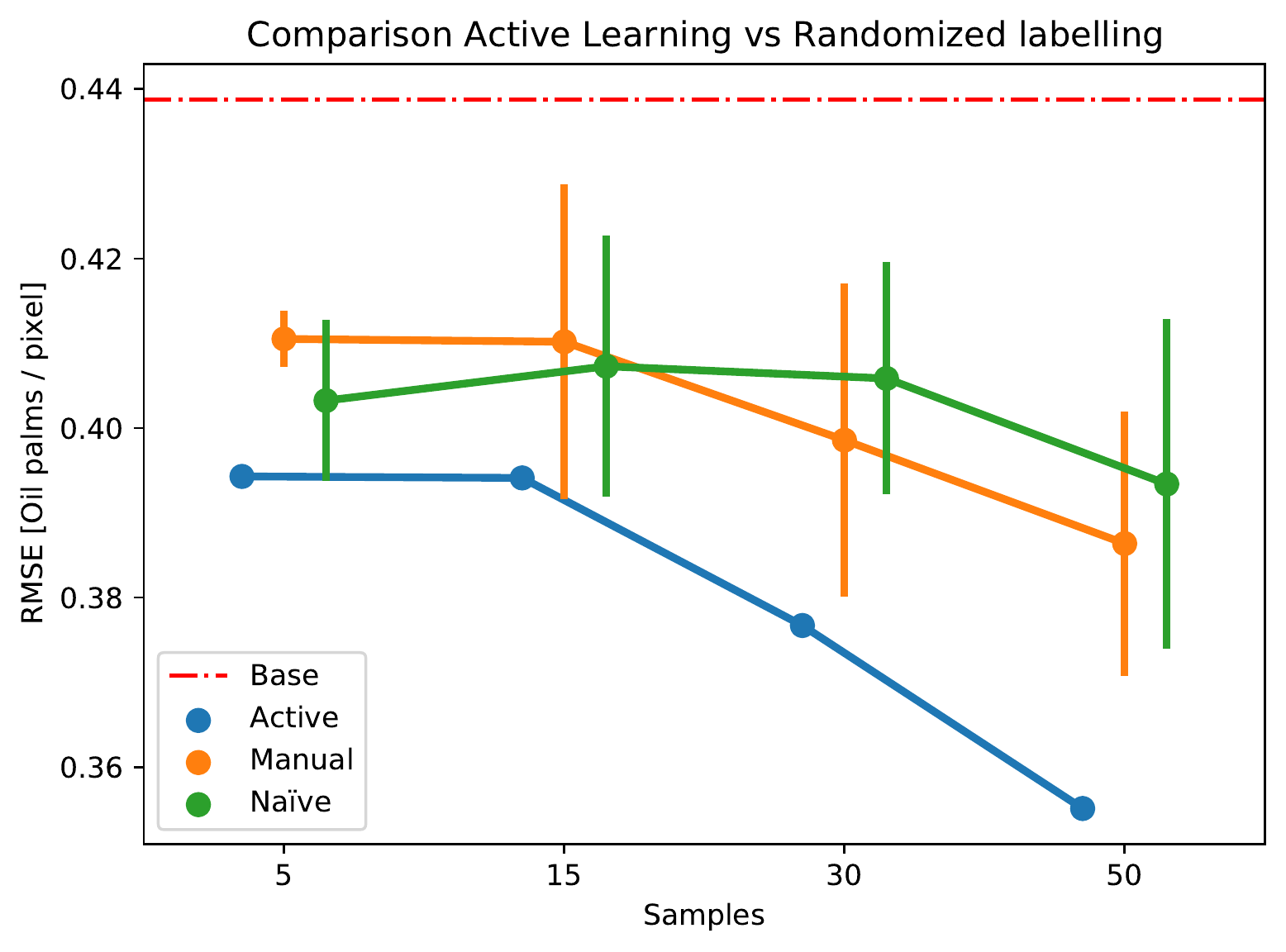}
    \caption{Comparison of active learning approach vs.\ two baselines. Error bars are standard deviations from 5 different runs of na\"{i}ve and manual selection.}
    \label{fig:results_fully}
\end{figure}

To start with, we \adda{defined} our interest region $\mathcal{X}$ made up of a very large number $N$ of samples $x$. \add{In our case, each individual 10m pixel from Sentinel-2 Images in the area of study is one sample $x$}.  
\add{We create\adda{d} an Active set $\mathcal{A}$ }
\remove{decide}
for which we annotate\adda{d} a subset of $x$ (i.e., Sentinel-2 pixel) with the desired outputs $y$ (i.e., oil palm density) using an annotation budget $B$, such that we can fit the model parameters $\theta$ and then apply the model to all the remaining (unlabelled) samples.
The goal of active learning is to
\remove{reduce the annotation effort by cleverly picking the right training data points, such that fewer of them are needed to capture all}
\add{cleverly pick the most informative training data points until exhausting the budget $B$, such that we capture the most }relevant correlations between radiometric input patterns in $x$ and output densities $y$.

\subsubsection*{\add{Base initial training $F_0(x)$}}
\remove{However,} This leads to a chicken-and-egg problem: in the absence of any information about the model \remove{$F_\theta$} \add{$F$ } we cannot measure how important a sample $x$ is for fitting it, except for the general intuition that the sample set should be diverse, so as to capture the variability of the inputs.
Hence, one has to annotate an (ideally, small) initial training set $\mathcal{L}=\{x^l,y^l\}$ and train a preliminary model \remove{$F_\theta^0$} \add{$F_0$}.
That function can now be applied to the set $\mathcal{U}=\{x^u\}$ of unlabelled samples to obtain estimates 
\remove{$\hat{y}^u=F_\theta^0(x^u)$}
\add{$\hat{y}^u=F_0(x^u)$}.
\subsubsection*{\add{Acquisition function}}
The fundamental assumption of active learning is that one can determine from $\hat{y}^u$ which samples should be labelled to improve the model, even without \add{having} access to the true values $y^u$ (and thus to the correctness of $\hat{y}^u$).
To that end the estimates are scored with an \emph{acquisition function} $g(\hat{y})$, and \add{the ones with the highest scores} \remove{the most high-scoring ones} \add{are used to construct $\mathcal{A} = \{x^*\}$, their corresponding labels $y^*$ are obtained with manual annotators, and a new model $F_1$ is obtained using $\mathcal{L}$ and the new samples $\{x^*,y^*\}$}
\remove{are annotated  and added to the training set for the new model $F_\theta^1$.}
The function $g(x)$ is typically based on the estimated \add{uncertainty of the prediction, often in the form of a predicted }variance $\sigma^2(\hat{y})$.\remove{of the prediction,} Following the reasoning that the model must be improved in those regions of the data space where it is uncertain~\cite{gal2017deep,haut2018active,Vit2020}.

Even if $g(\hat{y})$ is an effective proxy for the actual objective to decrease the prediction error, the described procedure is only optimal if new samples are added one by one until the available annotation budget $B$ is exhausted at iteration $F_\theta^B$.
Unfortunately, it is not practically viable to run the computationally expensive model fitting every time a sample has been added, so one has to add multiple samples at a time to the training set -- ideally one could add all $B$ samples in one shot and retrain only once.
This leads to a new problem: the set \add{$\mathcal{A}$ containing  the} $B$ highest scoring samples can (and often will) contain redundant patterns, i.e., the scores $g(\hat{y}_i),g(\hat{y}_j)$ are similarly high because their corresponding image observations $x_i,x_j$ are also similar.
But showing the same pattern to the model multiple times at the cost of excluding other, yet unseen ones is wasteful and will undermine the idea of active learning.

Consequently, the acquisition function\add{, in this case, }should be a set function that evaluates the benefit of adding an entire set of training data points, rather than \remove{an} individual points.

\subsubsection*{\add{Core-set approaches}}
\remove{
One way to do this is the \emph{core-set} approach.}
A core-set is a smaller subset that \emph{summarises} a large data collection in terms of the relevant properties for some task -- for instance in the case of clustering,
\add{ a good core-set is one that yields clusters similar to the ones obtained by clustering the complete (usually much larger) data set~\cite{bachem2018scalable}.}
\remove{it should yield similar clusters to those that would be found by the much more expensive clustering of the whole data set~\cite{bachem2018scalable}.}
In our case, a good core-set is a training set that leads to \add{predicted} \remove{regression} outputs similar to those that would be learned if annotations were available for all data points.
The use of core-sets for active learning was pioneered by~\cite{sener2017active}. However, their work has two major drawbacks: (\emph{i}) it ignores the model uncertainty; and (\emph{ii}) it does not scale to large data sets, as it requires an exhaustive set of pairwise (dis-)similarities between data points to assess diversity.
This second point highlights a more general principle: set functions are in general more expensive to compute than per-point scores, so we need to design our acquisition function carefully to ensure it remains tractable even for large values of $N$ and $B$.

\subsubsection*{\add{Proposed acquisition function}}
To address the two limitations \add{previously mentioned}, we propose to \add{construct $\mathcal{A}$ with samples to be labelled by} \remove{construct the core-set} using the following acquisition function:
\begin{equation}
g(x_i) =  \underbrace{\frac{ \sigma^2(\hat{y}_i) }{\sum\limits_{j\in \mathcal{U}} \sigma^2(\hat{y}_j)}}_\text{Uncertainty}+
  \underbrace{\frac{ d^z(x_i,\mu)^2 }{\sum\limits_{j\in\mathcal{U}} d^z(x_j,\mu)^2}}_\text{Diversity}\;,
\label{eq:sample_dist}
\end{equation}
where $\mu$ is the centre of gravity (mean value) of the \add{projected} data points $x_j\in\mathcal{U}$ \add{in a suitable feature space $z$}, and $d^z$ is \add{the corresponding} distance measure.
\remove{Typically $d^z$ is not computed directly in the input space of image intensities, but rather in some suitable feature space.}
\add{Instead of computing distances directly in the input space of image intensities, we compute them in a space $z$ that provides meaningful distances with respect to our task at hand.}
In our case the natural choice is to compute the (Euclidean) distances between deep activation maps in the deep network, see Section~\ref{sec:implementation} below.

The first term is simply the variance of the prediction (normalised over the unlabelled set) and corresponds to the standard assumption in active learning, that points with high uncertainty should be added to the training set.
The price to pay is that the $g(x)$ is no longer a true set function, and does not prevent the selected samples from lying far from the mean but close to each other in the data space.

\subsubsection*{\add{Active learning set $\mathcal{A}$ construction}}

Our solution to re-introduce pairwise dissimilarity is to select a larger set of $q$ candidate points (in our implementation, $q=1\cdot10^5$) with the highest scores $g(x_i)$, then cluster them into $B$ clusters with weighted $k$-means, assigning each sample the weight$\frac{1}{q\cdot g(x_i)}$. \add{To construct $\mathcal{A}$,} from each cluster we retain\adda{ed} only the candidate that lies nearest to the cluster centre.
This simple trick ensures that no two training samples are closer to each other than the cluster radius.
Note that, moreover, the clustering makes the selection more robust against outliers, which might otherwise dominate the core-set selection, as they tend to lie far from the mean $\mu$. \add{See Figure \ref{fig:method_overview} for an overview of our entire workflow.}

\begin{figure}[t]
    \centering
    \includegraphics[width=0.95\linewidth]{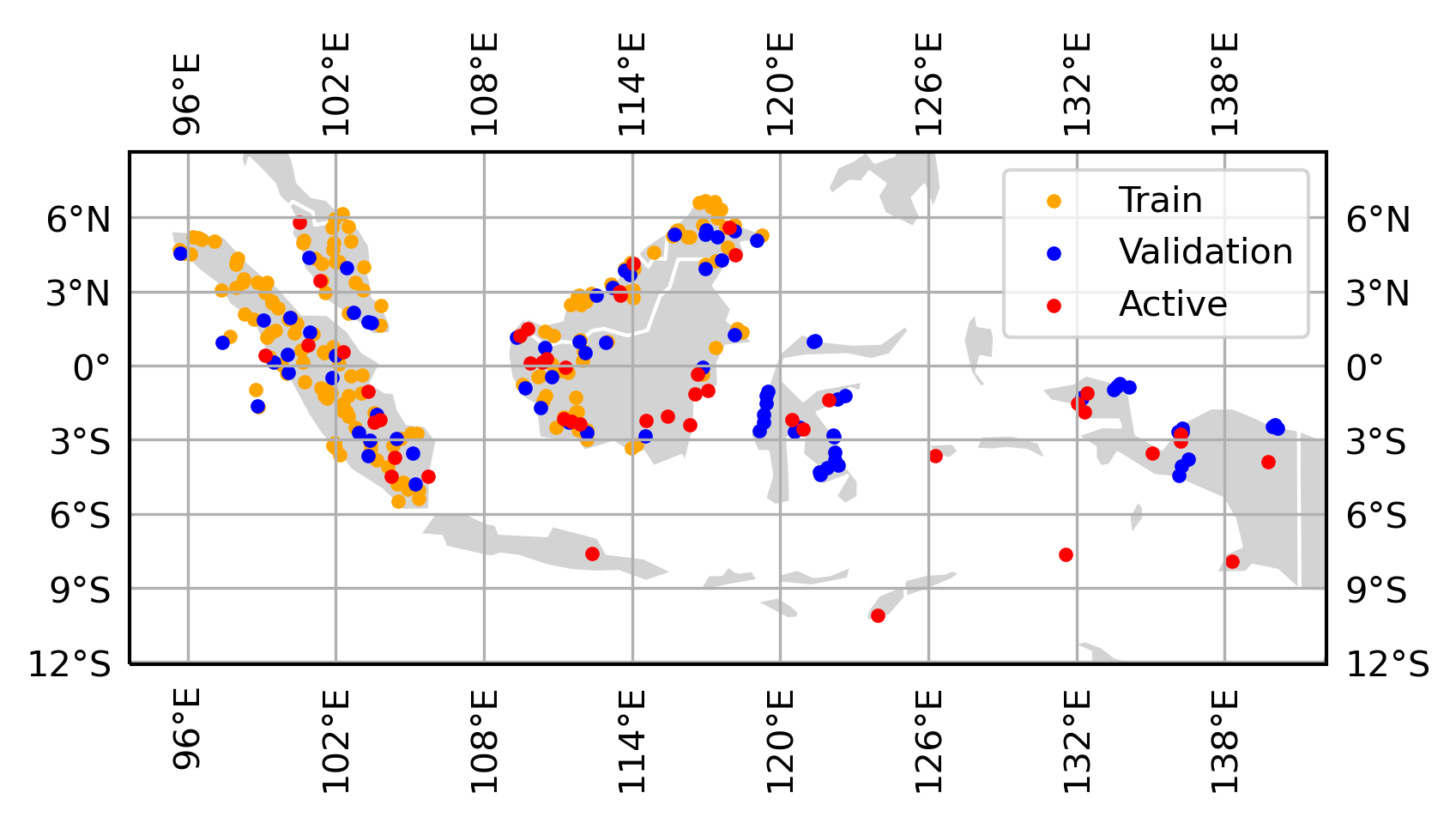}
    \caption{Full dataset with 30 active samples}
    \label{fig:AL_samples_completearea}
\end{figure}

\begin{figure}[h]
     \centering
     \includegraphics[width=0.8\linewidth]{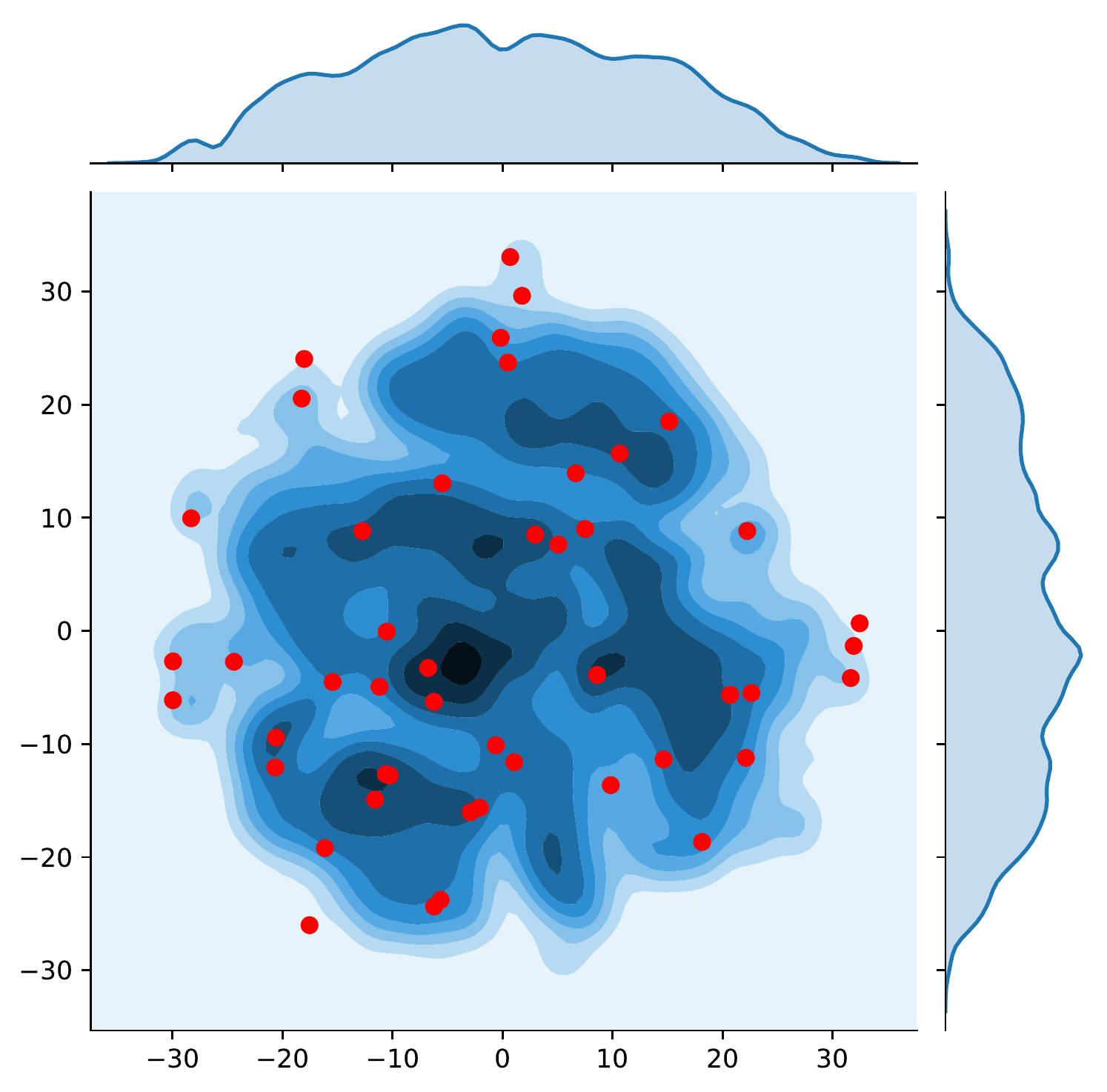}
     \caption{2-D tSNE visualisation of the deep features from the core-set selected for the area of study. In red the selected areas for labelling.}
    \label{fig:tsne_coresest}
 \end{figure}
 
\subsection{Implementation}\label{sec:implementation}

After introducing the general framework, we go on to discuss practical details of the technical implementation, with a particular focus on large-scale deployment with images totalling hundreds of giga-pixels.

\subsubsection*{Uncertainty Estimation}

To select informative samples, we must estimate the prediction uncertainty $\sigma^2(\hat{y})$ (first term in eq.\eqref{eq:sample_dist}).
More specifically, we are looking for samples with high \emph{epistemic} uncertainty~\cite{kendall2017uncertainties} \add{, i.e. the uncertainty that arises from the model parameters due to limited training data.} \remove{, indicating shortcomings of the (preliminary) model.}
A standard way to represent \remove{that}\add{epistemic} uncertainty numerically, if the model fitting includes stochastic elements, is to learn multiple instances of the model and view their predictions as Monte-Carlo samples from the posterior distribution of the output $\hat{y}$.
For the case of neural networks, random initialisation and stochastic gradient descent provide the basis for such an ensemble strategy.
A more efficient alternative is Monte-Carlo dropout~\cite{gal2016dropout}, where the model ensemble is approximated by multiple runs of the same model, where in each run a given proportion of the model parameters \remove{$\theta$} is suppressed. 
In this way one avoids having to train multiple models, at the cost of potentially less well-calibrated uncertainty estimates~\cite{ovadia2019can}.
With an ensemble of $T$ model predictions it is straight-forward to derive an ensemble prediction $\bar{y}=\frac{1}{T}\sum_{t=1}^T \hat{y}_t$ and and uncertainty estimate
\begin{equation}
\sigma^2(\hat{y}) = \frac{1}{T}\sum_{t=1}^T (\hat{y}_t-\bar{y})^2
\label{eq:uncertainty}
\end{equation}

\subsubsection*{Distance function and deep feature embedding}

To quantify the distance $d^z(x_i,x_j)$ between two samples we 
use\adda{d} the learned feature representation in the second-last layer of the neural network \add{$F$}.
To achieve a meaningful embedding that is representative across the complete area of study, we also include\adda{d} a pixel's location, such that the distance function can adapt to geographical variations. To inject the geographical location into the network we use\adda{d} $\mathsf{Space2Vec}$ encoding~\cite{Mai2020Multi-Scale}, which relies on a grid representation of relative and absolute positions and allows the model to learn correlations at different scales. 
\add{Intuitively, $\mathsf{Space2Vec}$ uses each pixel (longitude and latitude) location and computes its angle with respect to different directions in space, relative distances from each directions are computed at different scales, this allows our model to incorporate information from other pixels at different scales.
For more details refer to \cite{Mai2020Multi-Scale}.
}

Overall, the following transformation is applied to an input $x$ at location $(x_\text{lat,lon})$: 

\begin{align}
\text{location encoding:}\quad & r= \mathsf{Space2Vec}(x_\text{lat,lon}) \\
\text{CNN activations:}\quad & e= F^{e}(x)\\
\text{attention:}\quad & a= A([r,e]) \\
\text{final embedding:}\quad & F^{z}(x)= a \cdot [r,e]
\end{align}
where $[r,e]$ denotes concatenation along the channel dimension. $\mathsf{Space2Vec}$ is trained on our small, initial training set $\mathcal{L}$. The function \remove{$E$} \add{$F^{e}$} just reads out the activations \add{(i.e. the output a specific layer in the CNN)} from the density prediction network \remove{$F_\theta$} \add{$F$}. The simple attention mechanism $A$ consists of a $1\times1$ convolution followed by a sigmoid function and determines the relative importance (weight) of \remove{the spatial and image features} \add{the features from each pixel in the image}. The final embedding $F^{z}(x)$ is \add{the product of the CNN activations and the location encoding; since the latter is a value between $[0,1]$, it can be interpreted as a re-weighting value for CNN activations.} \remove{attention-weighted product of the location encoding and the CNN activations.}
The similarity between two data samples is then defined as the squared Euclidean distance between their embeddings, $d^z(x_i,x_j)=\|z_i-z_j\|_2^2$.

\subsection{Large scale implementation}\label{sec:large}

The second term of Eq.~\eqref{eq:sample_dist} is the distance from  a sample $x$ to the dataset mean $\mu$. 
Computing each individual distance $d^z(x,\mu)^2 $ as well as the cumulative distance can easily be parallelised across multiple machines for large-scale deployment.
To do this one simply splits the area of interest into $Q$ mutually exclusive regions and processes each of them on a different machine. 
We first compute\adda{d} the number of samples $n$, the sum $v$ and sum of squares $w$ of the deep embedding $z$, for all samples in $Q$.
Then we collect\adda{ed} all statistics from all regions and compute the global statistics $N = \sum_{Q} n$ and $\mu = \sum_{Q} v / N $.
Now, in parallel for each region, we can compute the average distance to the mean $d^z(x_q,\mu)= w - 2v\mu + N\mu^2$, and after collecting those distances from all regions we can compute the global distance to the mean and evaluate $g(x)$ for all the regions $Q$. For uncertainty the procedure is straightforward as, in order to compute the global uncertainty, we only require the cumulative sum over the uncertainties inside each region $Q$.
See Algorithm~\ref{alg:core-setconstruction} in the supplementary material for the pseudo-code of \add{ active learning method and Section \ref{sec:supl_details} for more details on the architecture}.

%% file: 3_experiments.tex
 \begin{figure}[t]
    \centering
    \includegraphics[width=\linewidth]{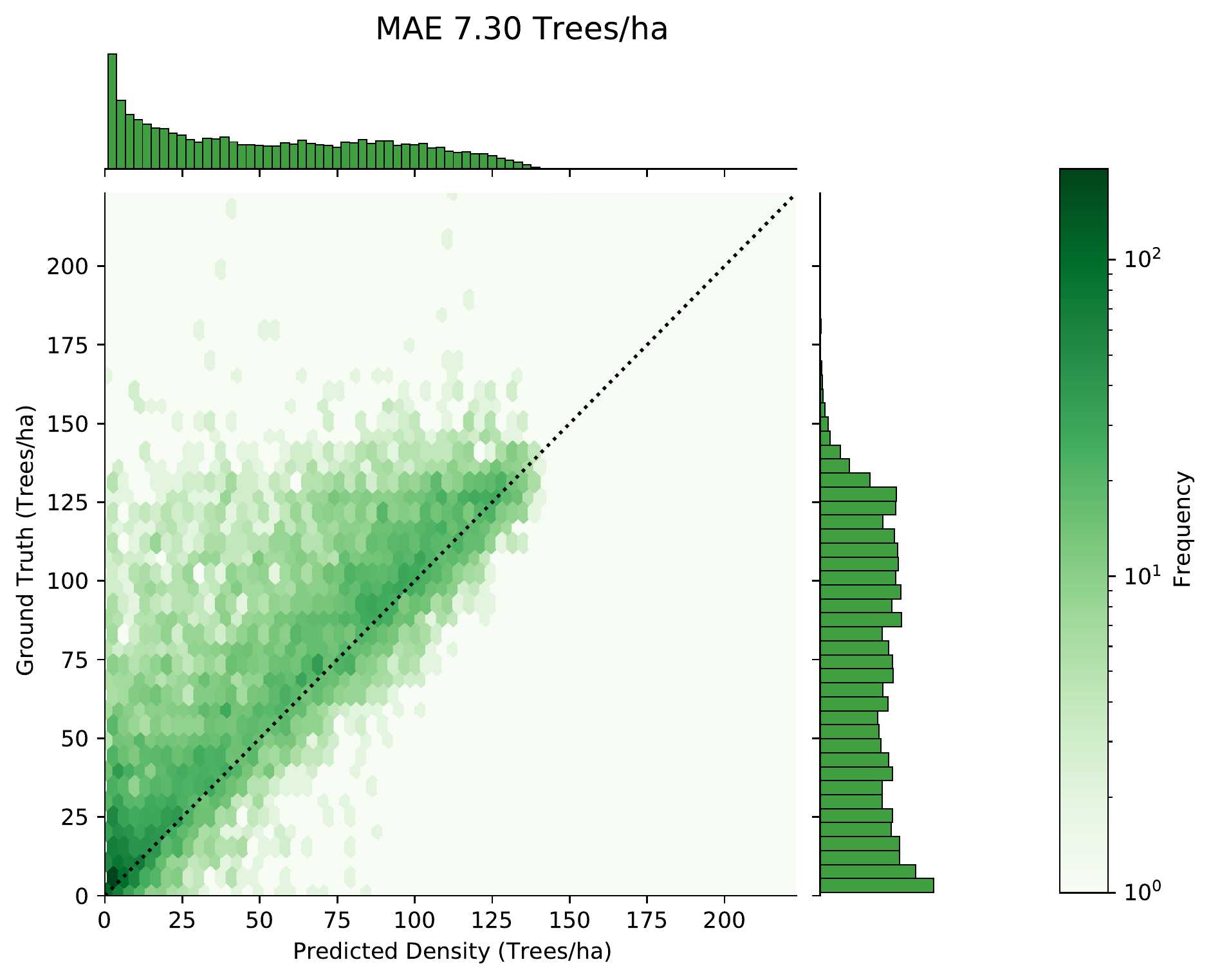}
    \caption{Predicted and true oil palm densities in the validation blocks (blocks with densities $<$1 were trimmed for visualisation). 
    }
    \label{fig:dens_validation_scatter}
\end{figure}

\begin{figure}[t]
    \centering
    \includegraphics[width=0.95\linewidth]{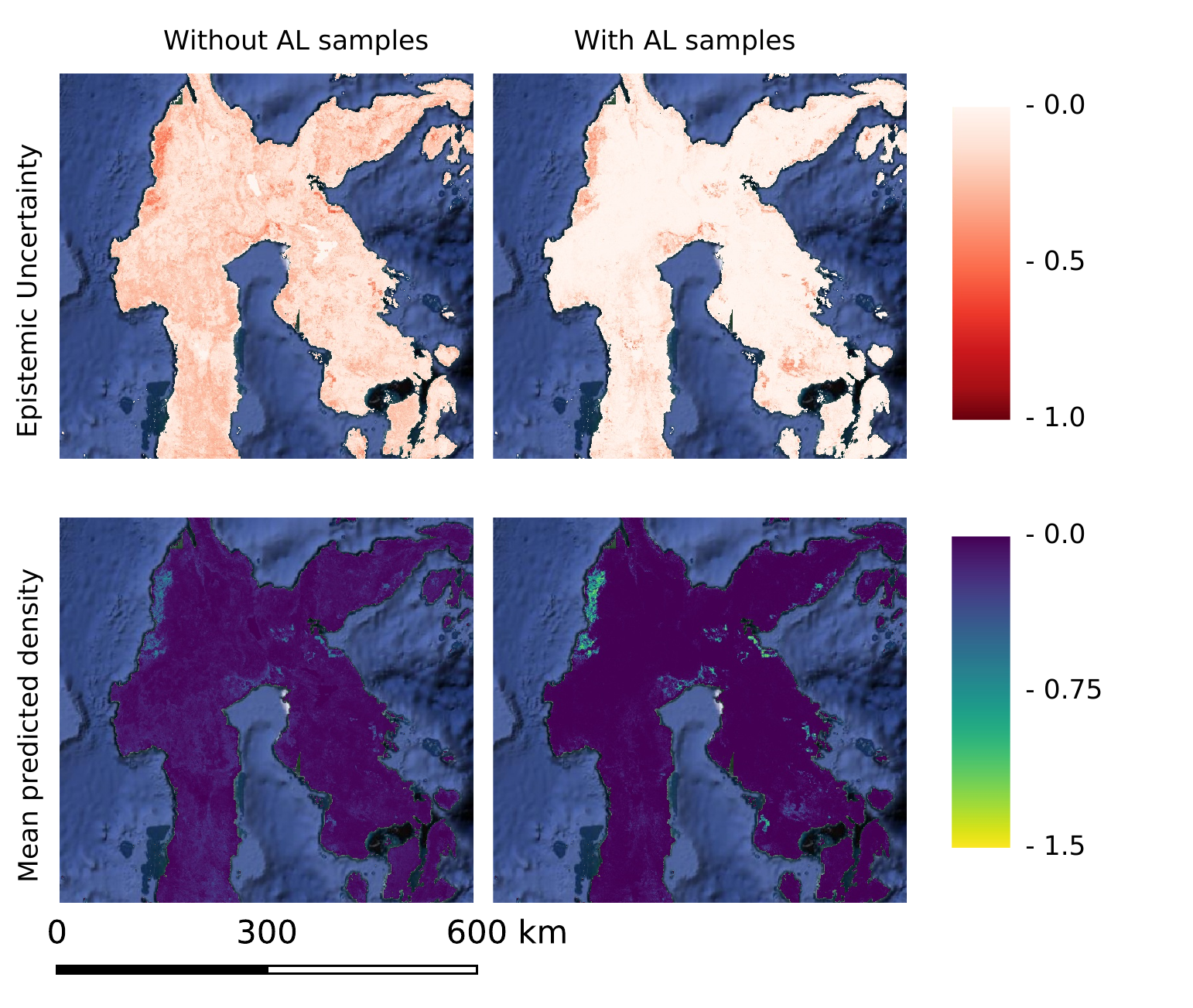}
    \caption{Comparison of uncertainty estimates after training with AL samples (\add{central} Sulawesi, Indonesia; aggregated to 500$\,$m for visualisation \add{)} }
    \label{fig:uncertainty_before_after}
\end{figure}

\begin{figure*}[t!]
    \centering
    \includegraphics[width=\linewidth]{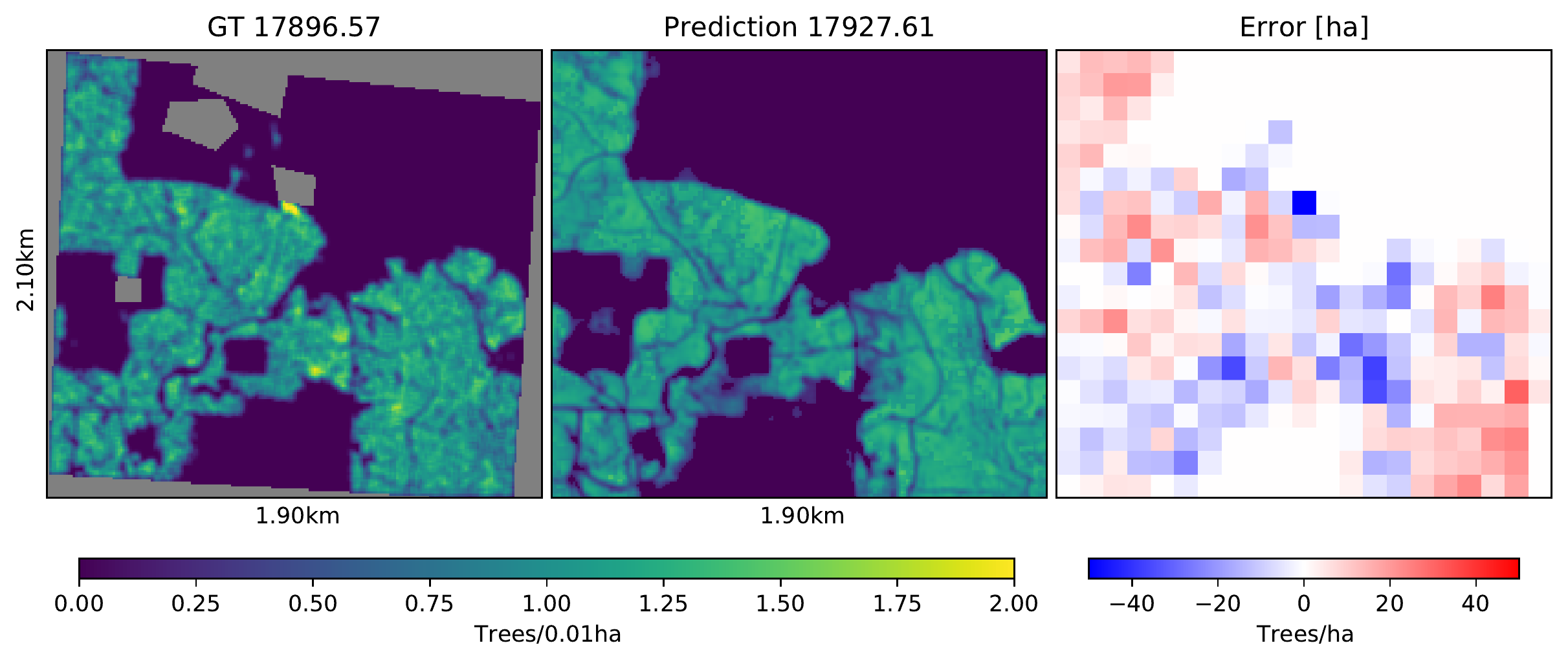}
    \includegraphics[width=\linewidth]{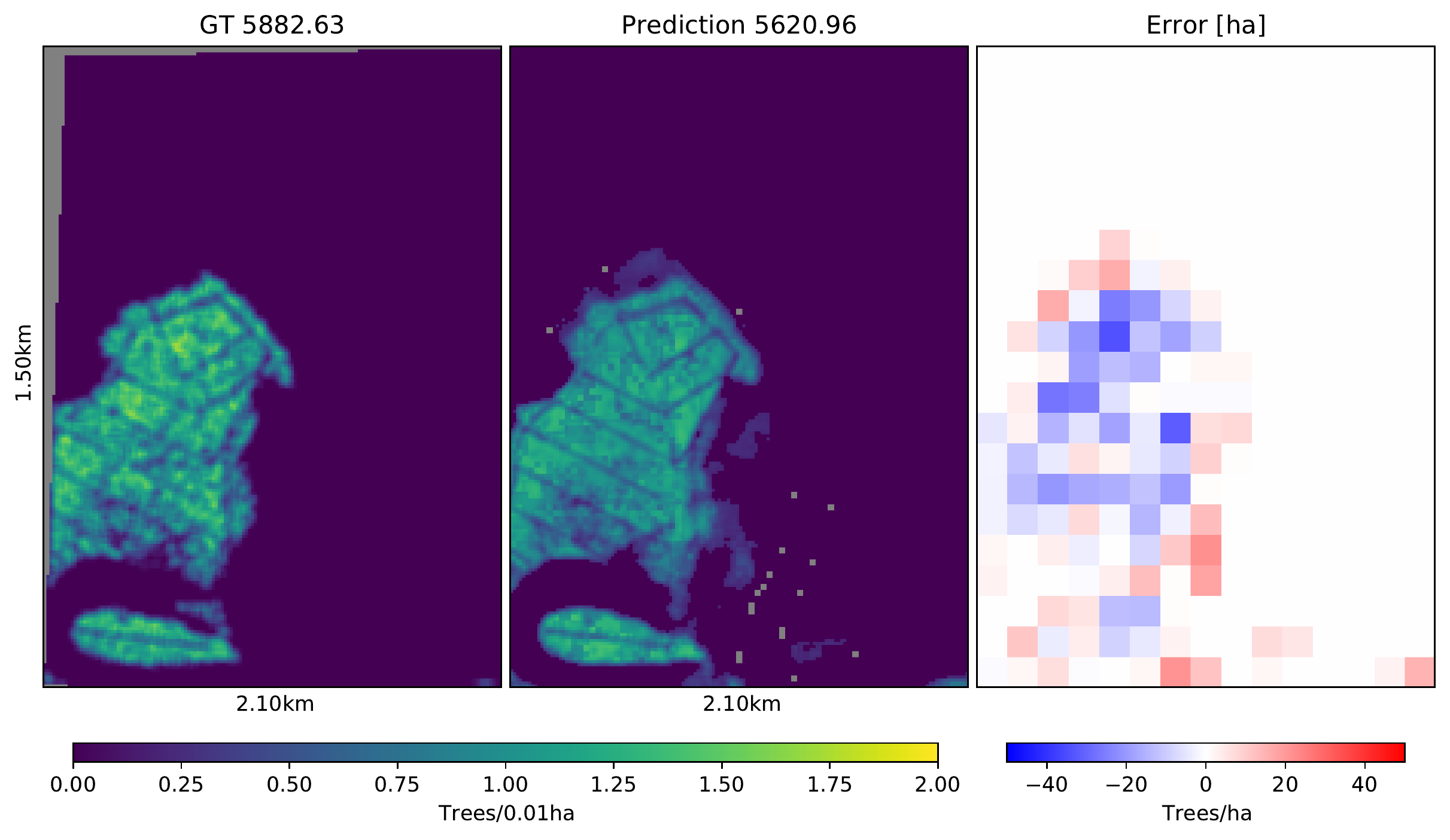}
    \caption{Qualitative Results density estimation with manually annotated data. Left: densities per 0.01$\,$ha block. Right: errors in Trees/ha}
    \label{fig:dens_validation_qual}
\end{figure*}

\begin{figure*}[t!]
\begin{subfigure}{0.495\linewidth}
  \centering
  \includegraphics[width=0.95\linewidth]{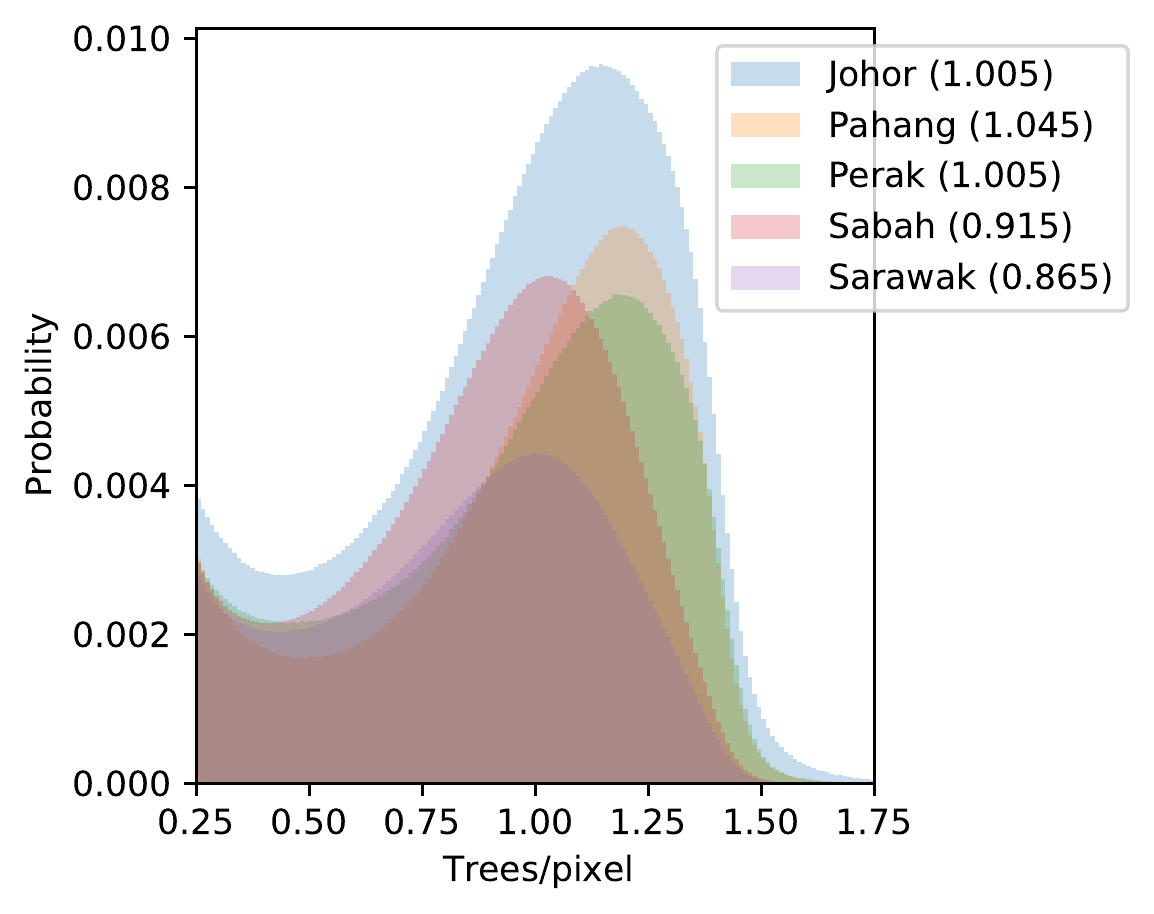}
\end{subfigure}
\begin{subfigure}{0.495\linewidth}
  \centering
  \includegraphics[width=\linewidth]{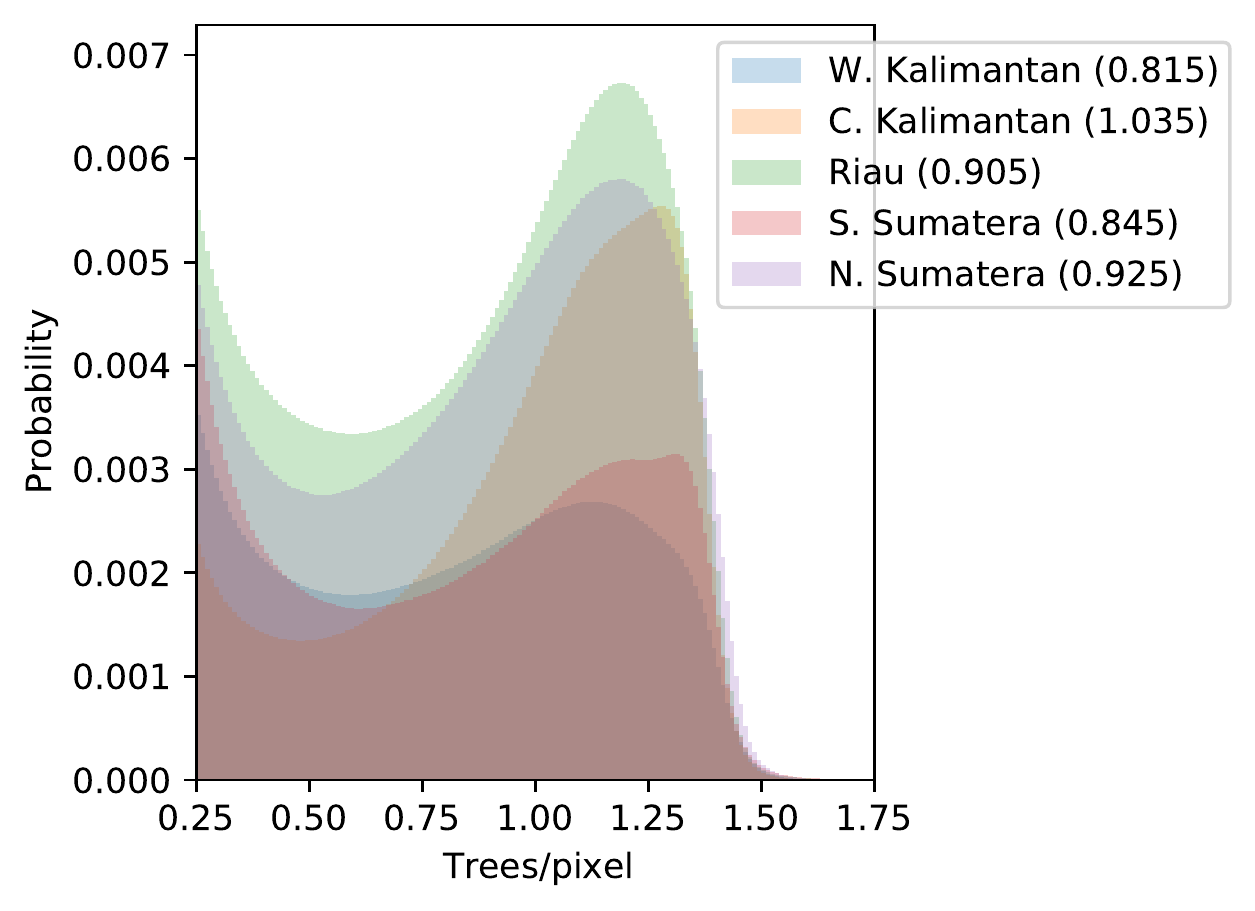}
\end{subfigure}
\caption{Histogram of pixel densities, for the five states with most oil palms in Malaysia and Indonesia for 2017}
    \label{fig:densities_bothcountries}
\end{figure*}

\begin{figure}[t!]
  \centering
  \includegraphics[width=\linewidth]{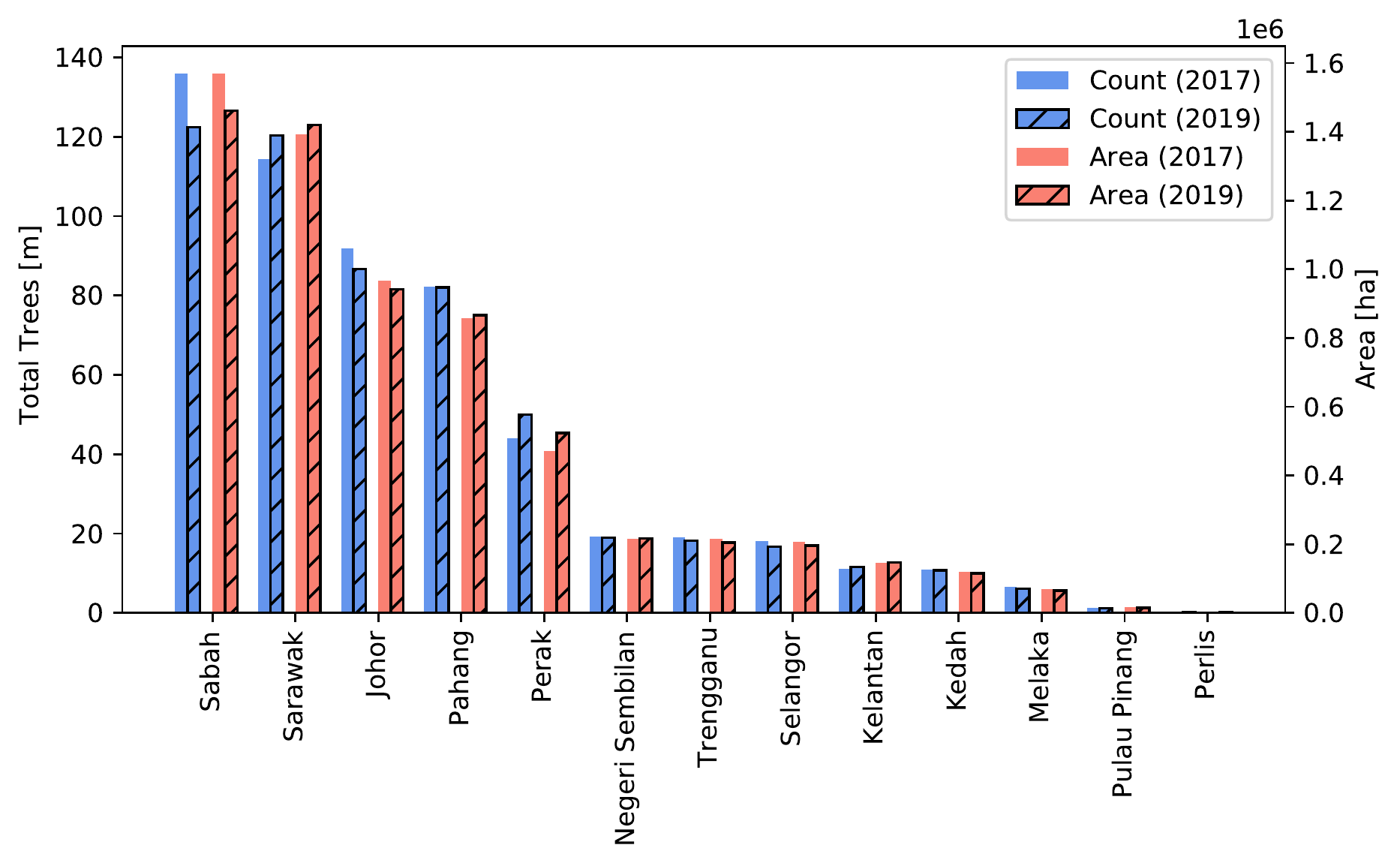}
\caption{Total number of oil palms vs.\ total planted area for the top-10 states in Malaysia for 2017 and 2019. Only pixels with density \textgreater0.2 were counted as oil palms.}
    \label{fig:counts2years_malaysia}
\end{figure}

\begin{figure}[t!]
  \centering
  \includegraphics[width=\linewidth]{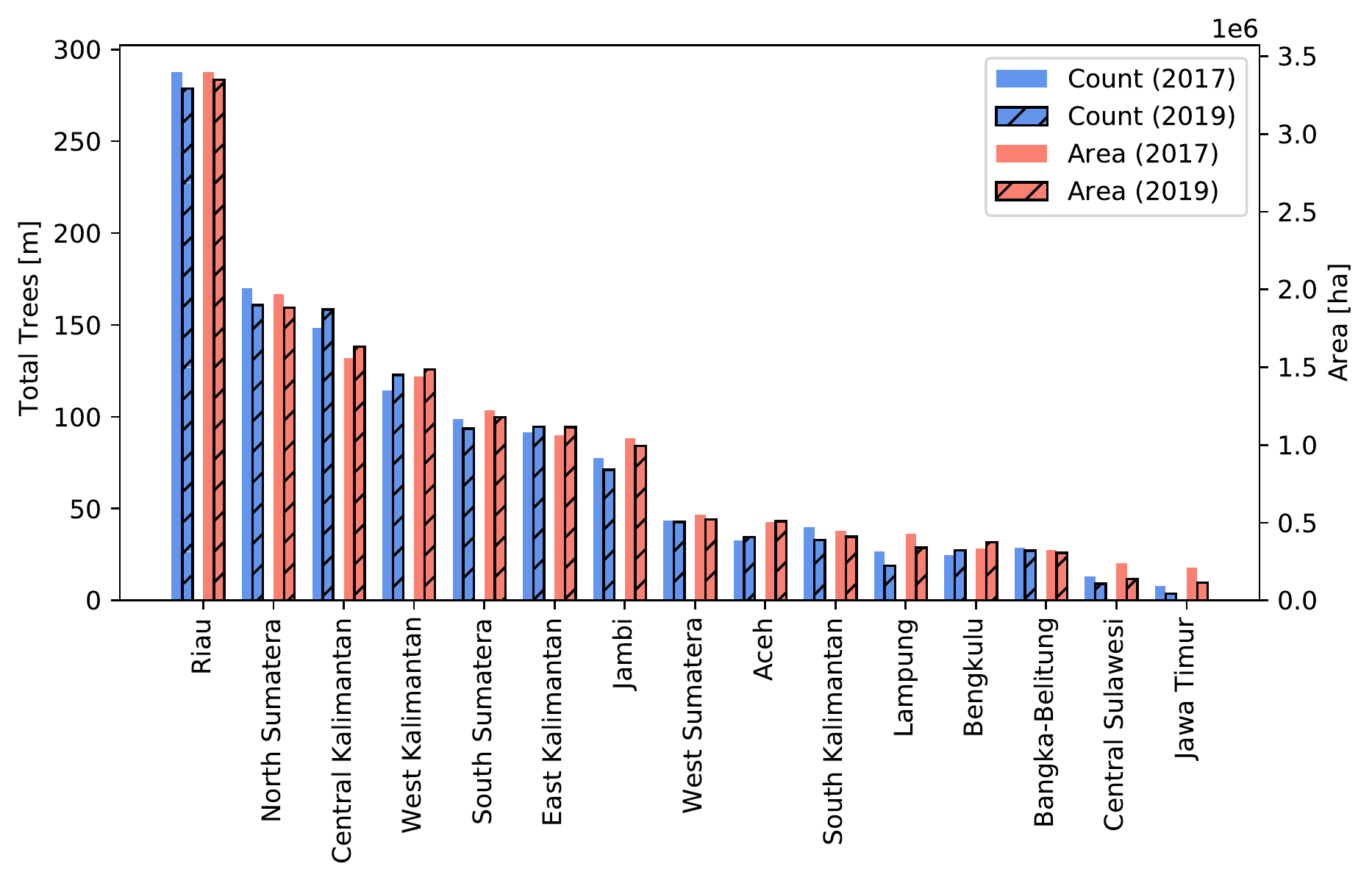}
\caption{Total number of oil palms vs.\ total planted area for the top-10 states in Indonesia for 2017 and 2019. Only pixels with density $>$0.2 were counted as oil palms.}
    \label{fig:counts2years_indonesia}
\end{figure}

\section{Experiments}\label{sec:experiments}

\subsection{\add{Active learning }reference dataset}
\label{sec:fullylabelled}

A common procedure to validate active learning methods is to take a fully annotated dataset and simulate the active learning procedure, by letting the acquisition function select new ``unlabelled" samples and reading out their labels.
However, that strategy is not viable for realistic, large-scale applications like ours. Labelling all oil palms in South-East Asia manually to benchmark our method is prohibitive. We thus restrict the validation experiments -- and thus the labelling effort -- to a smaller subset and label 126 blocks of at least 250$\,ha$ each in Sumatra, Indonesia and Peninsular Malaysia. Each block consists of approximately 160$\times$160 pixels of Sentinel-2 data. 
As a base dataset, we randomly sample 10 training blocks and 10 validation blocks. The ``unlabelled" dataset $\mathcal{U}$ is formed by the remaining $106$ blocks. As in this setup the unlabelled dataset $\mathcal{U}$ is not large, we skip the $k$-means clustering step and directly use the top-$k$ samples \add{with respect to the acquisition function $g(x)$ }during active learning.
For measuring the uncertainty of the estimates $\hat{y}$, we evaluated the quality of MC-dropout vs.\ an explicit ensemble. In line with other works, we find that uncertainty estimated with an explicit ensemble of 5 models showed a better calibration w.r.t.\ the true residuals in the validation areas. See Supplementary Material for details.

Using different sample sizes $B$, we evaluate how much active learning improves performance compared to the original base dataset, and compare it against two alternative approaches: 
\begin{description}
    \item[Na\"ive  selection:] here we add the same number of samples as with active learning, but in this case we pick random clusters of samples that are geographically close; thus simulating a na\"ive annotator.
    \item[Manual selection:] To manually sample the area of interest, an expert annotator should aim to cover as good as possible the area of interest in terms of visual diversity and geographical location. This is how the 106 manually selected available samples were constructed, so to simulate different number of samples, we uniformly choose from that manually curated set.
\end{description}
In both cases, we run the experiment with five different random seeds, to quantify how much variance is introduced by the randomness of the selection. Fig.~\ref{fig:labelled_example} illustrates an example of the selected samples for the different strategies. As it can be seen the three methods behave differently. The na\"ive and manual annotation strategies behave according to their design: the former selects points close to each other, whereas the latter annotates points spread across the entire region. Active learning results in a mixed strategy that aims for geographical spread, but avoids certain regions where none of the samples has high uncertainty. 

Figure~\ref{fig:results_fully} shows the estimation error of the three methods for different annotation budgets $B$. The manual and na\"ive annotation strategies lead to similar performance, with only slightly lower errors for manual selection, within the standard deviation of different runs. Active learning, on the other hand, consistently reduces the error, at all annotation budgets.
Note, no standard deviation can be given for AL, because in the closed world of the available 106 annotated samples the selection is deterministic, and the same points will be picked in each run.
We further note that the advantage of AL is smaller with very few samples and increases with the sample size $B$, a behaviour also observed in~\cite{sener2017active}.

\subsection{Large-scale mapping}
\label{sec:large_scale_mapping}
We start with a manually labelled base dataset of 250 regions (each with at least 250$\,ha$), out of which which we use 166 for training, and 84 for validation. After training a model on the base training dataset, we aim to predict and label 50 new active samples.
We define each region $q$ to have size $144\,$ha, which in Sentinel-2 pixels represents an image patch of 120$^\text{2}\,$\remove{px}\add{pixels}.
This results in 2 million different regions, after removing regions completely covered by water.
We calculate the global 
mean $\mu$ of \remove{the} the $z(x)$ embedding from all regions in the area of study,
to compute the acquisition function $g(x)$ for each region. 
The complete region, covering 48$^\circ$ longitude, is divided into strips of 12$^\circ$. For each strip we draw $10^5$ samples and use the $k$-means algorithm to find 50 cluster centres to be annotated as additional training samples.
The cluster centres are distributed across the 12$^\circ$ strips proportional to their land area.
The active learning samples were then manually annotated, resulting in the final dataset shown in Figure \ref{fig:AL_samples_completearea}.
To visualise how diverse our chosen samples are compared to the rest of the dataset, we construct a t-SNE visualisation \cite{maaten2008visualizing}. %
t-SNE non-linearly projects high-dimensional vectors (in our case the deep features $F^z(x)$ from each sample) down to a low-dimensional vector (in our case 2-D vectors) while minimally distorting their distances.
Figure~\ref{fig:tsne_coresest} shows the t-SNE visualisation of our core-set samples. The actively selected samples (shown as red points) are well distributed across all deep features (blue), which indicates a high diversity among the selected samples.
\add{Furthermore, note that the selected samples are not only on oil palm plantations as the model may also be uncertain about other types of vegetation. We therefore added many samples on forest or crops that do not contain any oil palm trees.}
In practice, we ended up with 46 labelled areas; this difference arose because our method chose samples in areas where no cloud-free high-resolution images were available for labelling.

\begin{figure*}[t]
    \centering
    \includegraphics[width=\linewidth]{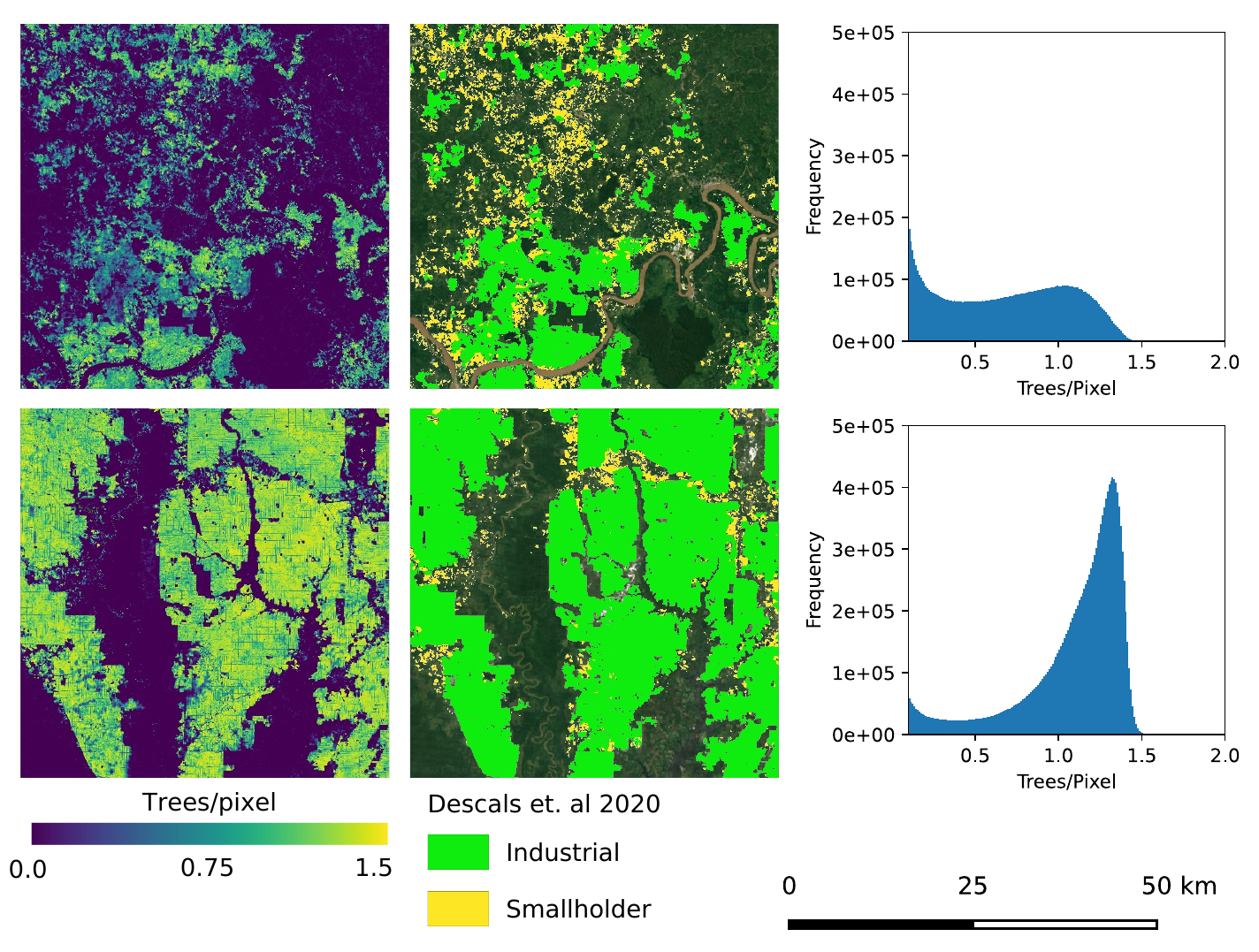}
    \caption{Predicted oil palm densities in smallholder and large-scale plantations}
    \label{fig:densities_small_large}
\end{figure*}

\subsection{Quantitative validation of oil palm densities}

We quantitatively evaluate our oil palm density estimates using validation regions not shown to the model during training. All reference data was obtained by manually annotating individual trees in very high resolution overhead images (GSD \textless30cm). At the Sentinel-2 pixel size of 10$\,$m the number of trees per pixel is low (generally \textless2), and the definition uncertainty of plantation boundaries is larger than the pixel size.
Thus, we aggregate both ground truth counts and predict densities into tree counts per 10$\times$10 pixel block (1$\,ha$) for the validation\footnote{Note that in Section \ref{sec:fullylabelled}, to avoid computational overhead for the different experiments, we evaluated directly on a 10~$m$ pixel level in contrast to a 1$\,ha$ level.}. Evaluating our error on 1$\,ha$ blocks allows us to compare if the method manages to retrieve the density for meaningful numbers of trees and avoid random fluctuations at pixel level.
As shown in Fig.~\ref{fig:dens_validation_scatter}, the Mean Absolute Error (MAE) over all validation sites is 7.30~trees/$ha$.
For comparison, using only the base training dataset, the MAE was 10.14~trees/$ha$, in other words the error without AL samples would be 39\% higher.
Fig.~\ref{fig:uncertainty_before_after} illustrates how the additional samples improve the prediction and reduce its uncertainty, in particular suppressing spurious low-density predictions in regions without oil palms.

Fig.~\ref{fig:dens_validation_qual} shows the true and predicted densities for two example blocks, and the corresponding errors at $ha$ level. Although some extreme density values are underestimated, the overall counts and structure of the plantations are retrieved with high correctness.
Since it is not feasible to manually label several billion individual oil palm trees of all Malaysia and Indonesia, the evaluation of oil palm density maps beyond the manually labelled validation samples has to remain limited to visual inspection and a qualitative analysis. 
We find that extending our initial training set of 166 blocks with just 46 additional blocks selected by AL greatly improves the map (Fig.~\ref{fig:uncertainty_before_after}). 

\subsection{Analysis of oil palm densities}
\label{sec:density_analysis}

To the best of our knowledge, we provide the first map of high-resolution oil palm densities across the major planting regions of South-East Asia. According to our estimate, there were $0.55\cdot10^9$ oil palms in Malaysia and $1.28\cdot10^9$ oil palms in Indonesia in 2017. Assuming a cut-off threshold of \textgreater0.2 trees/pixel for \remove{deliberately} planted crop areas, we estimate the total area of palm oil plantations in 2017 to be $6.24\cdot10^6\,ha$ in Malaysia, respectively $16.18\cdot10^6\,ha$ in Indonesia. For 2019, we estimate $0.54\cdot10^9$ oil palms covering $6.17\cdot10^{6}\,ha$ in Malaysia and $1.23\cdot10^9$ oil palms covering $15.29\cdot10^{6}\,ha$ in Indonesia.

Per-pixel tree counts allow us to evaluate how the tree density varies across different locations, or in function of other geographical factors. We show densities for individual states of Malaysia and Indonesia in Fig.~\ref{fig:densities_bothcountries} for 2017 and total estimated oil palms and covered areas for Malaysia in Fig.~\ref{fig:counts2years_malaysia} and Indonesia in Fig.~\ref{fig:counts2years_indonesia}.

\paragraph{Smallholder vs.\ large-scale plantations}

By combining our 2019 oil palm density map with the classification of \cite{descals2020hiresglobalmap} from the same year, we can compare the density distributions in smallholder plots versus large-scale plantations. By the definition of~\cite{descals2020hiresglobalmap}, smallholder oil palm plantations are smaller than 25$\,ha$, have heterogeneous tree age, and are less structured in terms of shape and layout. Some of these features correlate with density. In Fig.~\ref{fig:densities_small_large} (top row), we can see industrial and smallholder plantations as mapped by~\cite{descals2020hiresglobalmap}. Our findings support those empirical findings of~\cite{descals2020hiresglobalmap}: smallholder plantations exhibit lower tree densities and strong, local density variations.
Nonetheless, distinguishing smallholder plantations from large-scale plantations is difficult from Sentinel-2 satellite imagery. For example, there is no clear threshold to distinguish smallholder from industrial plantations based solely on density, because tree density inside industrial plantations can vary significantly, too (Fig.~\ref{fig:densities_small_large} (bottom row)).
On our maps one can also see density prediction highlights detailed structures inside individual plantations. In industrial plantations, for example, much lower densities are observed on the access roads between blocks of oil palms.

\subsection{Comparison between 2017 and 2019 oil palm density maps}
\label{sec:comparison}

\begin{figure}[!t]
    \centering
    \begin{subfigure}{\linewidth}
  \centering
    \includegraphics[width=\linewidth]{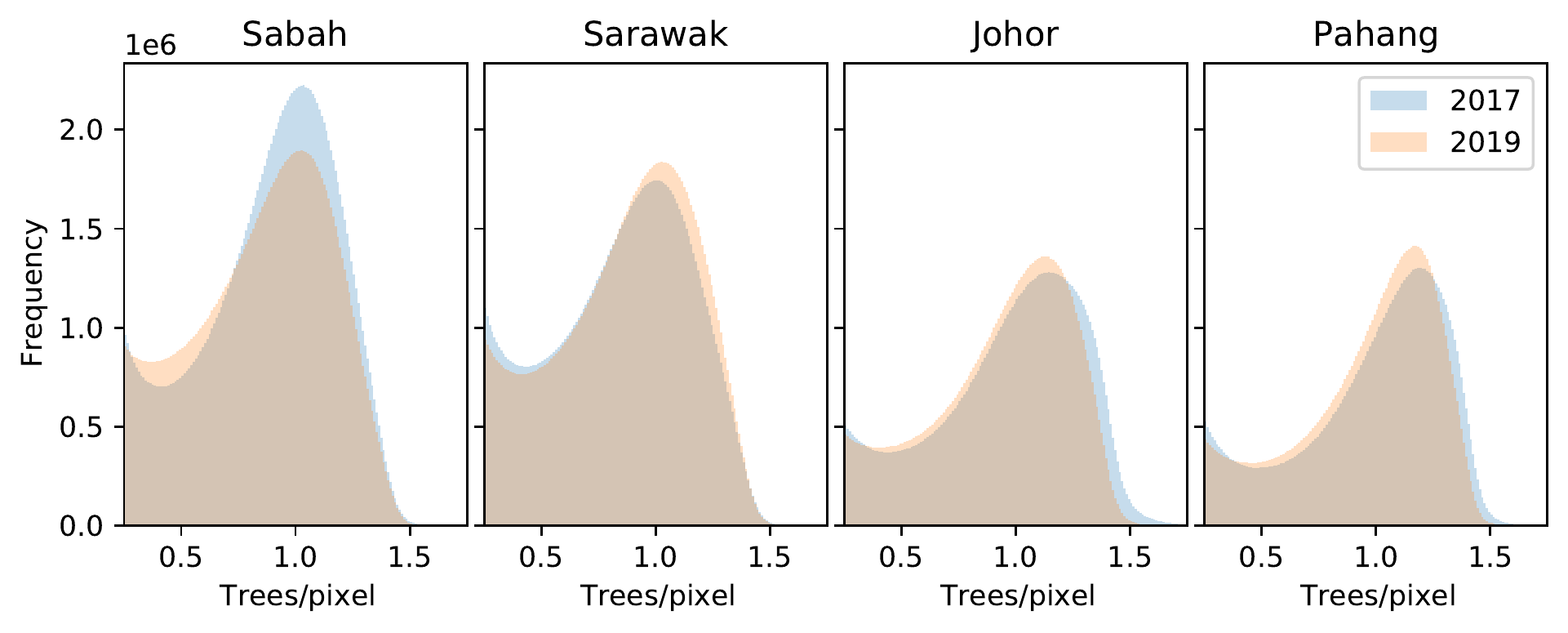}
  \caption{Malaysia}
  \label{fig:changes_dens_malaysia}
\end{subfigure}
\begin{subfigure}{\linewidth}
  \centering
    \includegraphics[width=\linewidth]{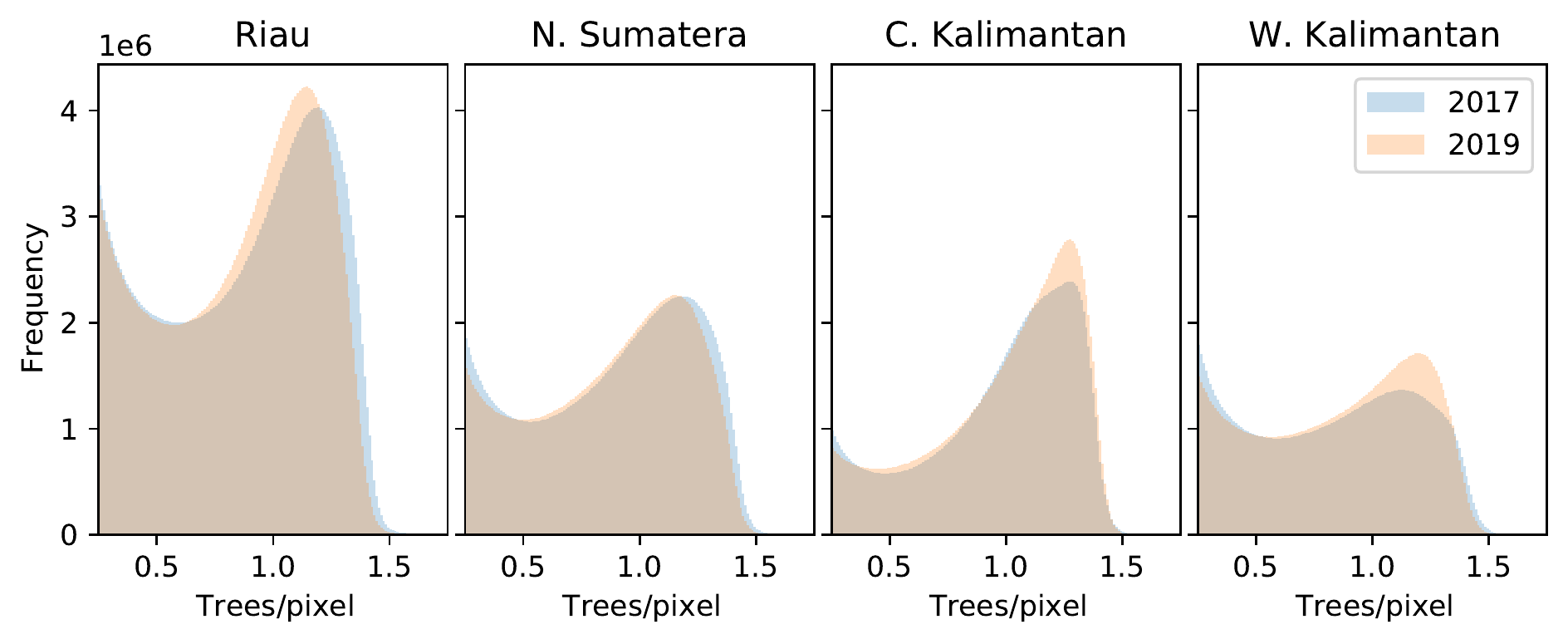}
  \caption{Indonesia}
\end{subfigure}
    \caption{Changes of oil palm density from 2017 vs 2019 in top-4 states by total area covered.}
    \label{fig:changes_density}
\end{figure}

\begin{figure}[!t]
    \centering
    \begin{subfigure}{\linewidth}
  \centering
    \includegraphics[width=0.49\linewidth]{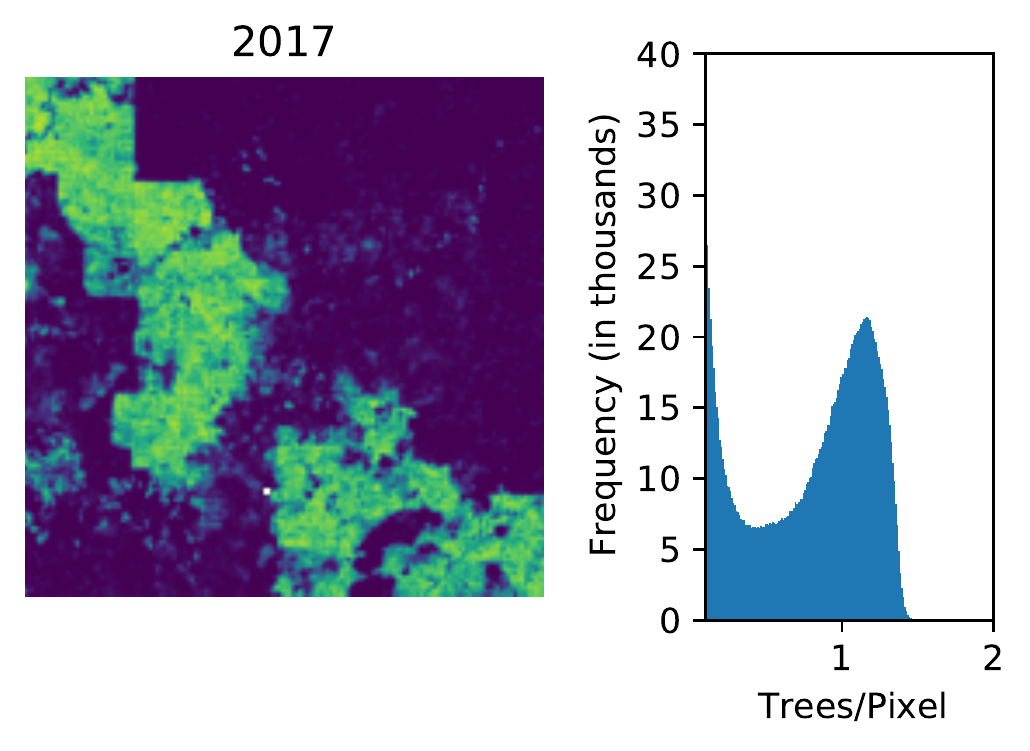}
    \includegraphics[width=0.49\linewidth]{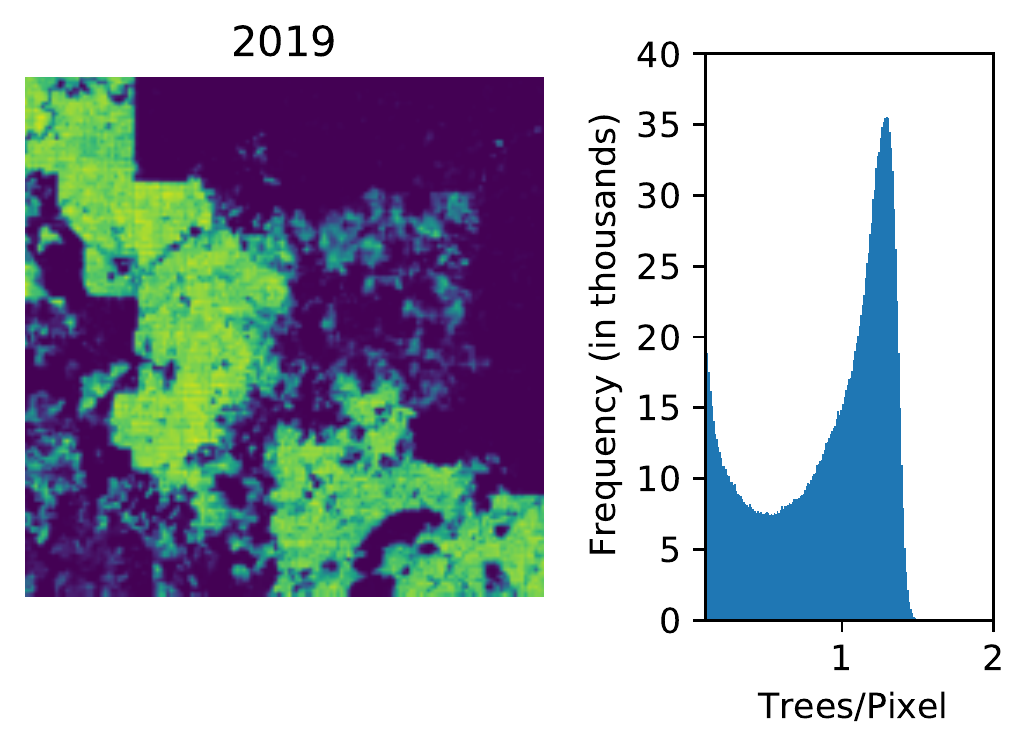}
       \caption{Central Kalimantan, Indonesia}
  \label{fig:changes_dens_detail_kalim}
\end{subfigure}
\begin{subfigure}{\linewidth}
  \centering
    \includegraphics[width=0.49\linewidth]{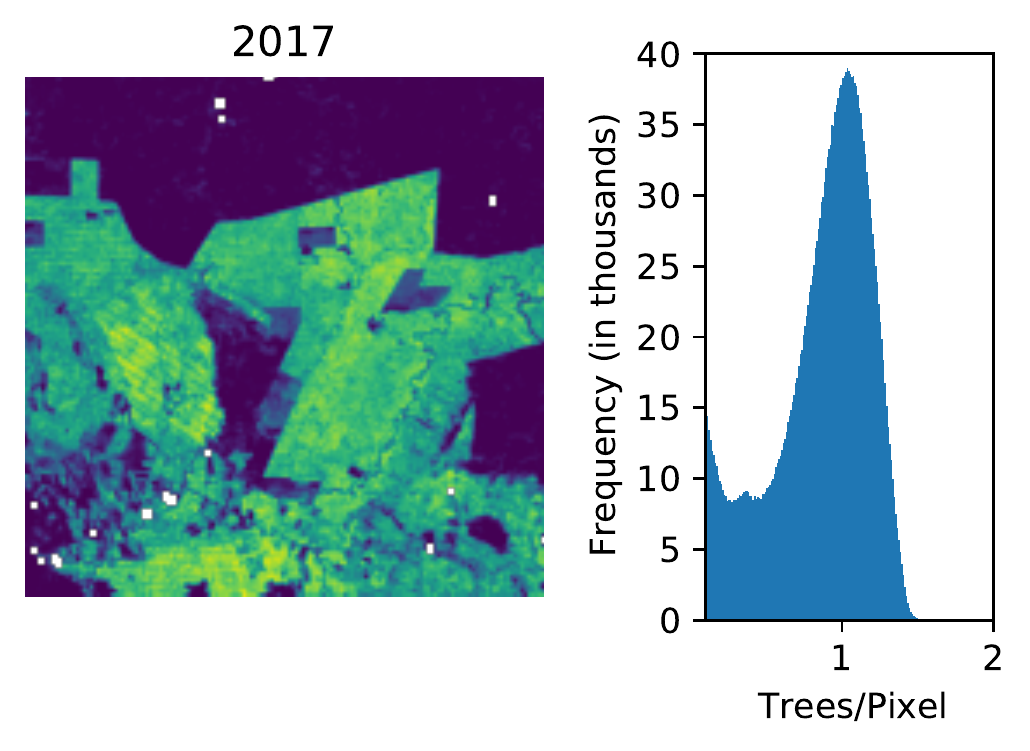}
        \includegraphics[width=0.49\linewidth]{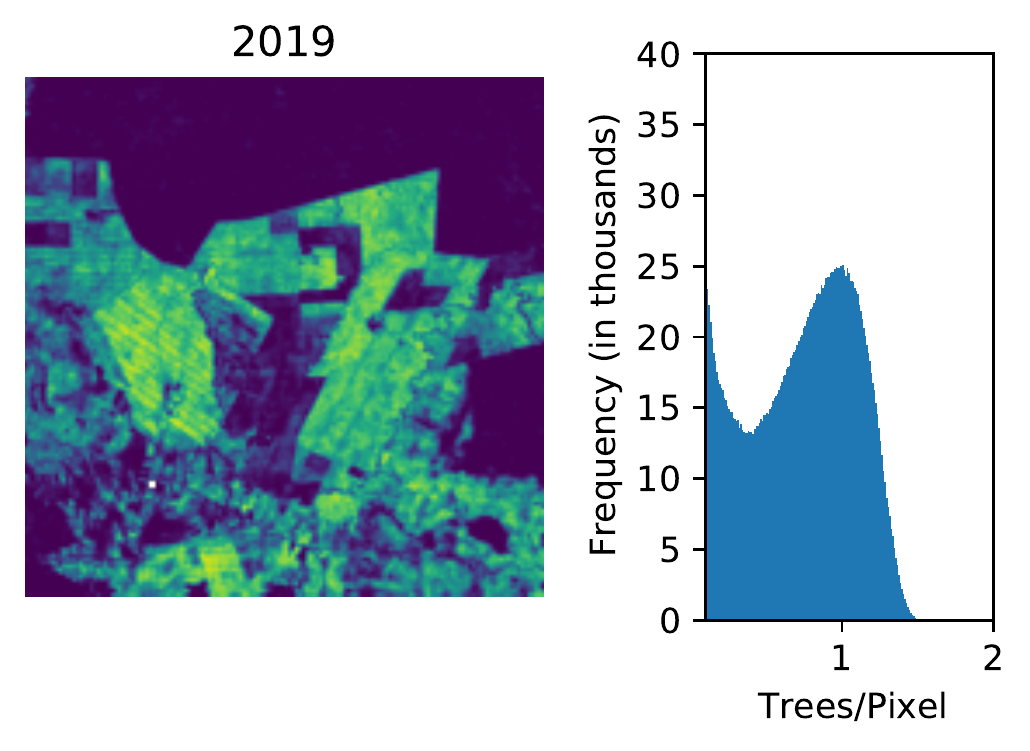}
    \caption{Sabah, Malaysia}
      \label{fig:changes_dens_detail_sabah}
\end{subfigure}
  \caption{Detail of changes of oil palm density from 2017 vs 2019 in 
  \add{selected areas}
  }
  \label{fig:change_dens_detail}
\end{figure}

We did not observe systematic density shifts from 2017 to 2019, which indicates that our model generalises well across different years.
This allows us to evaluate density changes between 2017 and 2019 as shown in  Fig.~\ref{fig:changes_density}. In Central Kalimantan, for example, the median increased from $1.035$ to $1.055$, which is most likely caused by new plantings with higher densities, as can be observed in the histogram. A detailed view of an example area with a large density shift is displayed in Fig.~\ref{fig:changes_dens_detail_kalim}, where higher density areas appear to correspond mostly to \remove{what in 2017 were young oil palm plantations} \add{new plantations in 2019}. \add{Furthermore, in Malaysia some areas showed a decline in the amount of planted oil palms, corresponding mostly to replanting schemes (Fig.~\ref{fig:changes_dens_detail_sabah}).} 

\subsection{Validation of predicted land-cover}

\begin{figure*}[hbtp!]
    \centering
    \includegraphics[width=0.8\textwidth]{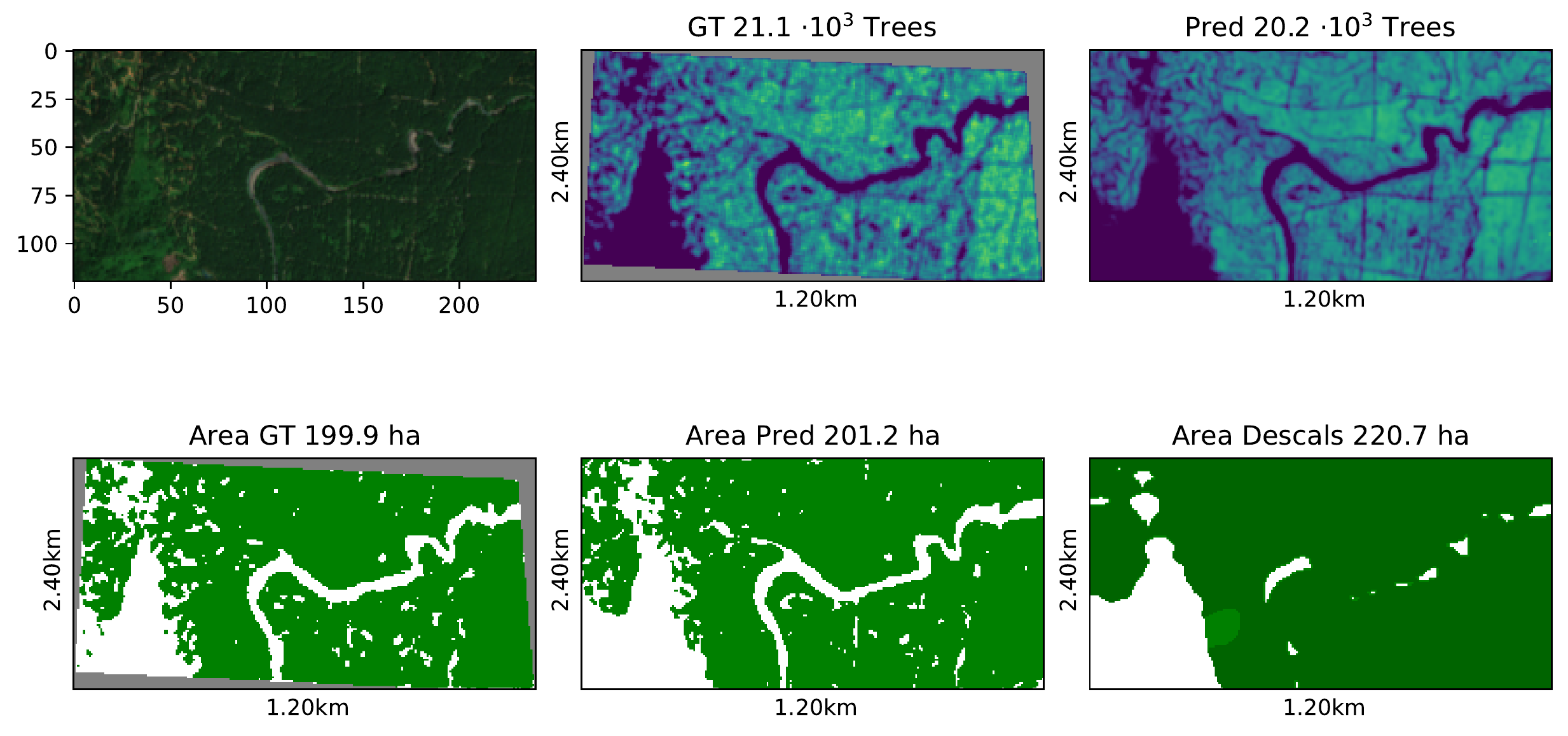}
    \includegraphics[width=0.8\textwidth]{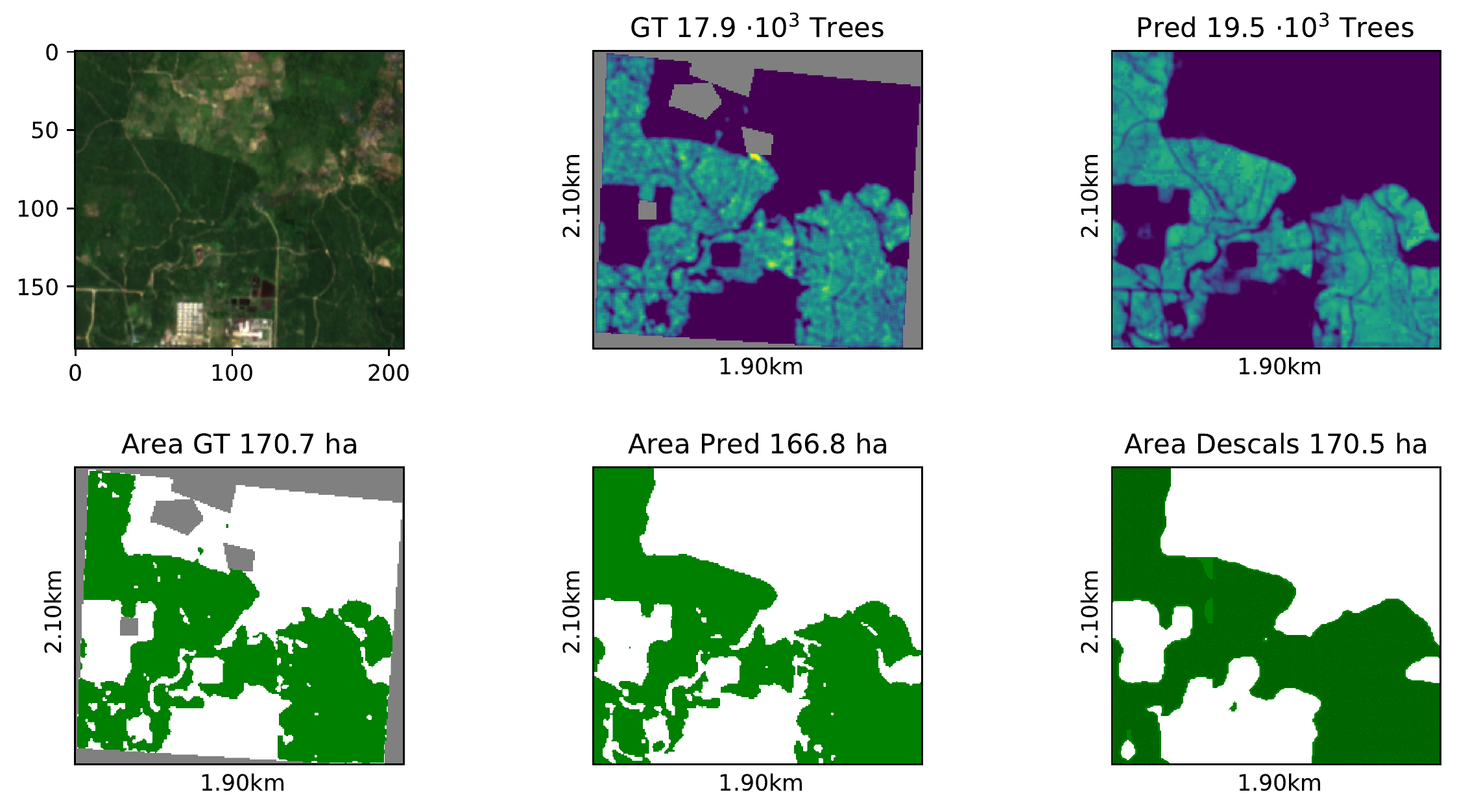}
    \includegraphics[width=0.8\textwidth]{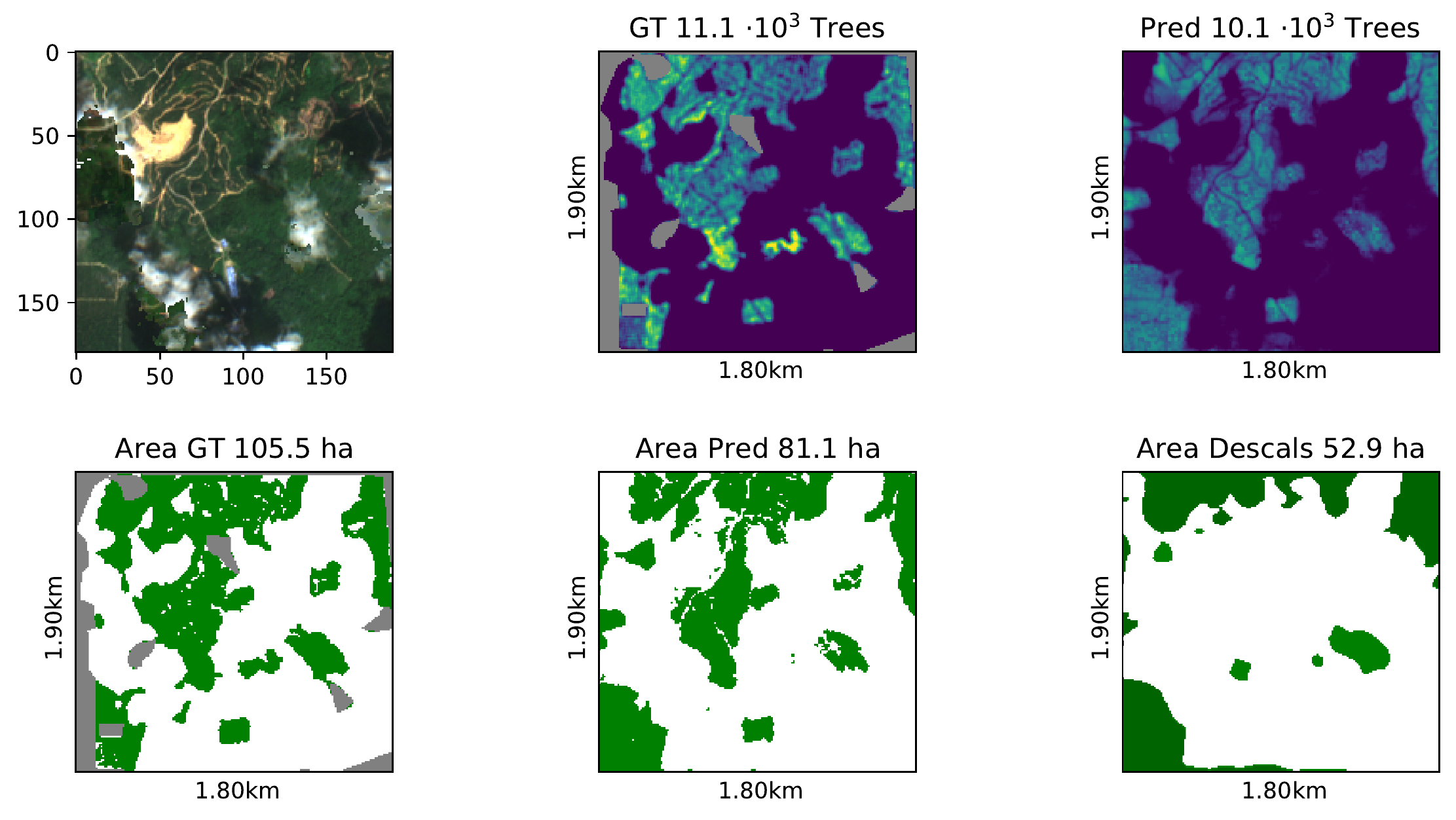}
    \caption{Comparison with \cite{descals2019oil} for 2019 selected areas. In Descals maps, Dark and light green denote industrial and small-holder plantations respectively}
    \label{fig:semantic_comparsion_descals}
\end{figure*}

Although our primary goal is density estimation, we can also compare our approach with land cover maps that only show binary presence/absence of oil palms. Recently \cite{descals2020hiresglobalmap} published a worldwide  map that focuses on classifying smallholder vs.\ industrial plantations. See Figure \ref{fig:semantic_comparsion_descals} for a comparison in selected areas.

\begin{figure}[h!]
    \centering
    \begin{subfigure}{0.49\linewidth}
      \centering
    \includegraphics[width=\linewidth]{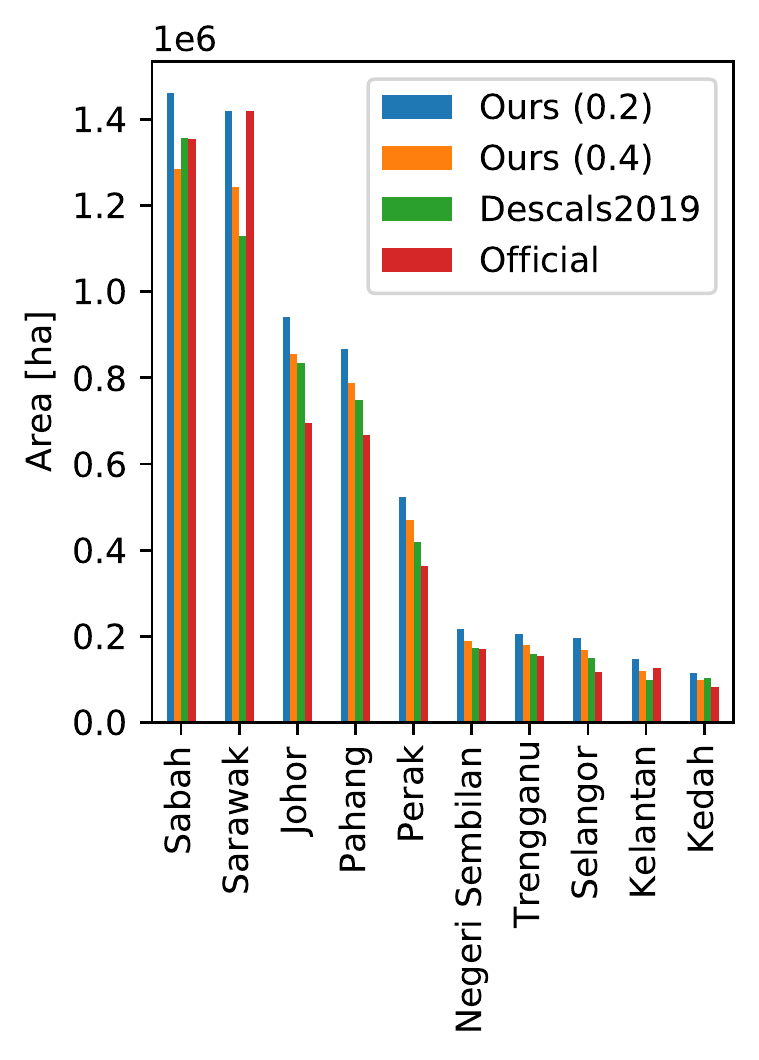}
    \caption{Malaysia}
    \end{subfigure}
    \begin{subfigure}{0.49\linewidth}
      \centering
    \includegraphics[width=\linewidth]{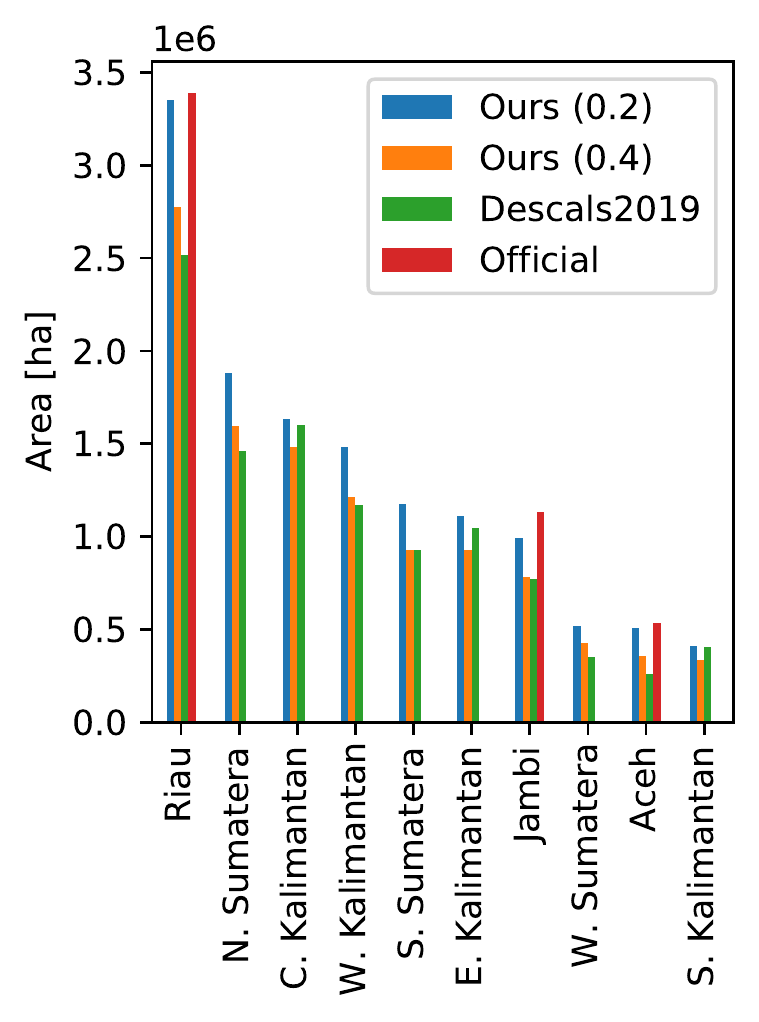}
    \caption{Indonesia}
    \end{subfigure}
    \caption{Predicted area of oil palm per state/country for 2019. Ours($t$) indicates estimated area with different minimum density thresholds per pixel $t\in \{0.2,0.4\}$. Official statistics from Malaysia and Indonesia are from MPOB\cite{mpob} and the Indonesian Ministry of Agriculture \cite{indo_agr}, respectively}
    \label{fig:predicted_area}
\end{figure}

The Malaysian Palm Oil Board (MPOB)~\cite{mpob} provides monthly statistics about the area planted with mature oil palms in every state. In Figure \ref{fig:predicted_area} we compare their estimates,
\cite{descals2020hiresglobalmap} and ours.
For Indonesia, we obtain official reports for 2019 from the Ministry of Agriculture \cite{indo_agr}.
We compare in Table \ref{tab:areas_malaysia} those to area estimates from our map, with cut-off thresholds of $\textgreater$0.2 and \textgreater0.4 trees/pixel. The lower threshold yields estimates closer to the official statistics in Indonesia, where the area has apparently been over-estimated, both \remove{w.r.t.} \add{with respect to} our estimates and to \cite{descals2020hiresglobalmap}. On the other hand, thresholding our estimates at 0.4 aligns better with \cite{descals2020hiresglobalmap}. These large differences in Indonesia could possibly be explained by the way the \remove{plantationa}\add{plantations} are delineated for the official statistics.

Overall, we observed a slight reduction of total planted area of $1.15\%$ for the whole country. \add{For example, in Sabah, Malaysia we found a decrease of 6.95\% (in contrast to official statistics reporting a decrease of only 1.9\%). As illustrated in Fig.~\ref{fig:changes_dens_detail_sabah} changes can be due to replanting schemes on previously densely planted areas.}
\begin{table}[th]
  \begin{center}
  	\resizebox{\columnwidth}{!}{%
    \input{tables/stats_per_state_malaysia}}
    \caption{Predicted area of oil palm ($10^3 ha$) per state in Malaysia, in 2017 and 2019. Official statistics taken from MPOB\cite{mpob}}
      \label{tab:areas_malaysia}
  \end{center}
\end{table}

%% file: tables/stats_per_state_malaysia.tex
\begin{tabular}{lrrrrrrr}
\toprule
{} & \multicolumn{3}{c}{Ours (0.2)}   && \multicolumn{3}{c}{Official}  \\
\cline{2-4}  \cline{6-8} 
  State &   \multicolumn{1}{c}{2017} &  \multicolumn{1}{c}{2019}  & $\Delta\%$ &&     \multicolumn{1}{c}{2017} &  \multicolumn{1}{c}{2019}  & $\Delta\%$ \\
\cline{1-8}\rule{0pt}{3ex} 
Sabah           &    1569.88 & 1460.79 &       -6.95 &&  1380.04 & 1353.81 &  -1.90 \\
Sarawak         &    1391.65 & 1419.54 &        2.00 &&  1342.10 & 1419.30 &   5.75 \\
Johor           &     965.78 &  941.99 &       -2.46 &&  682.62 &  694.10 &   1.68 \\
Pahang          &     857.21 &  866.48 &        1.08 &&   641.88 &  668.24 &   4.11 \\
Perak           &     470.62 &  523.39 &       11.21 &&   360.50 &  363.81 &   0.92 \\
Negeri Sembilan &     216.32 &  216.77 &        0.21 &&   162.63 &  170.97 &   5.13 \\
Trengganu       &     215.72 &  205.22 &       -4.87 &&   146.56 &  153.66 &   4.84 \\
Selangor        &     206.25 &  195.80 &       -5.07 &&   128.06 &  117.56 &  -8.20 \\
Kelantan        &     144.90 &  147.30 &        1.66 &&   118.09 &  127.22 &   7.73 \\
Kedah           &     118.75 &  115.25 &       -2.95 &&    82.42 &   81.79 &  -0.76 \\
Melaka          &      68.57 &   64.88 &       -5.38 &&    52.32 &   52.08 &  -0.46 \\
Pulau Pinang    &      16.13 &   14.44 &      -10.49 &&    12.87 &   13.45 &   4.47 \\
Perlis          &       4.72 &    2.84 &      -39.86 &&     0.62 &    0.84 &  36.47 \\
\cline{1-8}   \rule{0pt}{3ex} 
Malaysia  &    6246.50 & 6174.68 &       -1.15 &&  5110.71 & 5216.82 &   2.08 \\
\bottomrule
\end{tabular}

%% file: 4_conclusion.tex
\add{
\section{Discussion}\label{sec:discussion}

Our experiments indicate that oil palm density  can be efficiently retrieved at large scale from Sentinel-2 imagery. 
Per-pixel tree density has a number of potential advantages over conventional presence/absence land cover maps.
One advantage of density maps, compared to conventional land cover maps, is additional evidence about changes in palm tree density across different geographical regions (Fig.~\ref{fig:densities_bothcountries}). Such variations in density distributions can largely be attributed to different plantation patterns. For instance, in the Malaysian state of Sarawak, we observe the lowest median density among the compared states; this translates to a larger planted area for the same number of trees. According to official statistics of Fresh Fruit Bunch (FFB) yield in Malaysia\footnote{Figures for the total FFB yield of 2017, according to~\cite{mpob}}, the yield in Sarawak was $16.12\,t/ha$, compared to a national yield of 17.90, and values of 18.34 for Sabah, 20.64 for Johor and 17.91 for Pahang, respectively.
\add{We hypothesise that a lower tree density could have an impact on the lower yield per area in Sarawak. However, this is a complex process that requires long term studies to asses how exactly it is influencing the yield \cite{bonneau2018optimum,rafii2013variation}}.
}

Our work presented in this paper is limited to Malaysia and Indonesia, which produce over 80\% of global palm oil. In order to inform at global level, we need to expand to further growing regions like West Africa. 
We are currently using multi-spectral satellite images from the two Sentinel-2 satellites as data source. The major disadvantage of optical sensors is that they cannot see through the frequent cloud cover in tropical regions. We circumnavigate this shortcoming by computing density across stacks of Sentinel-2 images per location, in the hope that any point on the ground will be visible at least once per year. However, this incurs a substantial computational burden. Like~\cite{descals2020hiresglobalmap} we will thus explore the possibility to add Sentinel-1 SAR imagery to our approach, to be more robust against dense cloud cover. 

In Indonesia, official estimates were almost always higher than our estimates and those from \cite{descals2020hiresglobalmap}. In fact, different cut-off thresholds between oil palms and background for our method yield different estimates that range between the two sources. This could be an indication that the methodology used to report the official planted area differs from what we define as oil palm areas. For example, at a density threshold of 0.4, roads inside large-scale plantations would be excluded in our map, which is almost certainly not the case for the official statistics and for \cite{descals2020hiresglobalmap}.

We will make all 10~$m$ resolution maps for 2017 and 2019 available
upon publication of this paper.

\section{Conclusions}\label{sec:conclusion}

We have proposed an active deep learning approach with a computationally lightweight acquisition function $g(x)$ that selects large datasets for labelling very efficiently. Our active learning strategy allows to iteratively chose an optimal set of samples for interactive labelling, striking a balance between high diversity and high uncertainty of the selected samples. Approach and algorithm are designed to scale to entire world regions with billions of individual object instances. 
We hope to have shown that active (deep) learning is an excellent tool for remote sensing applications to environmental problems at very large scale, where manual annotation of sufficiently large ground truth is prohibitive.

We have applied our AL method to compute the first dense, 10-meter resolution map of oil palm densities of the world's two main producers, Malaysia and Indonesia, with a Mean Absolute Error of $\pm$7.30 trees/$ha$. 
According to our maps, in 2019 there were 0.54 billion oil palms covering $6.17\times10^{6}~ha$ in Malaysia and 1.23 billion oil palms covering $15.29\times10^{6}~ha$ in Indonesia.

%% file: 5_supplementary.tex
\section{Supplementary}
	\subsection{Implementation and Architecture Details}
	\label{sec:supl_details}	

\subsubsection*{Pseudo-code for Active Learning in Large scale datasets}
\label{sec:pseudo_code}

See Algorithm \ref{alg:core-setconstruction} for details on the large scale implementation of our proposed method.

\input{algorithm_1}

\add{
\subsubsection*{Architecture details}
	
	We implement all our experiments in Tensorflow using Python, see Tab.~\ref{tab:architecture} for the specific architecture details of our network $F$. For training we use a patch size of $16\times16$ pixels, a batch size of 128, trained for 100 epochs and used ADAM optimiser~\footnote{Kingma D.P., Ba J., Adam: A method for stochastic optimization Bengio Y., LeCun Y., 3rd International Conference on Learning Representations, ICLR 2015, San Diego, CA, USA, May 7–9, 2015, Conference Track Proceedings 2015} with a learning rate of $1\cdot10^{-4}$. From all labeled areas we randomly extracted a total of $1\cdot10^6$ patches proportional to the area of each labeled area.
	
	\begin{table}[t]
		\begin{center}
			\resizebox{\columnwidth}{!}{%
			\input{tables/architecture}}
			\caption{\textbf{Architecture $F(x)$ used in our method.} $x$ represents an input image from Sentinel-2, the output $\hat{y}$ is the estimated density for the corresponding input image $x$, $y_{cl}$ is the output of the auxiliary classification task, used only for training. Conv2D($n,k$) is a convolutional operations with $n$ filters and a filter of size $k$, before every Conv2D($n,k$) a BatchNorm Layer is applied \cite{ioffe2015batch}. Rectified Linear Units (ReLU) has the form $x_{\text{out}} = max(0, x_{\text{in}}) $ }
			\label{tab:architecture}
		\end{center}
	\end{table}
}

\subsection{Quality of uncertainty estimates}

\begin{figure}[t]
    \centering
    \includegraphics[width=0.95\linewidth]{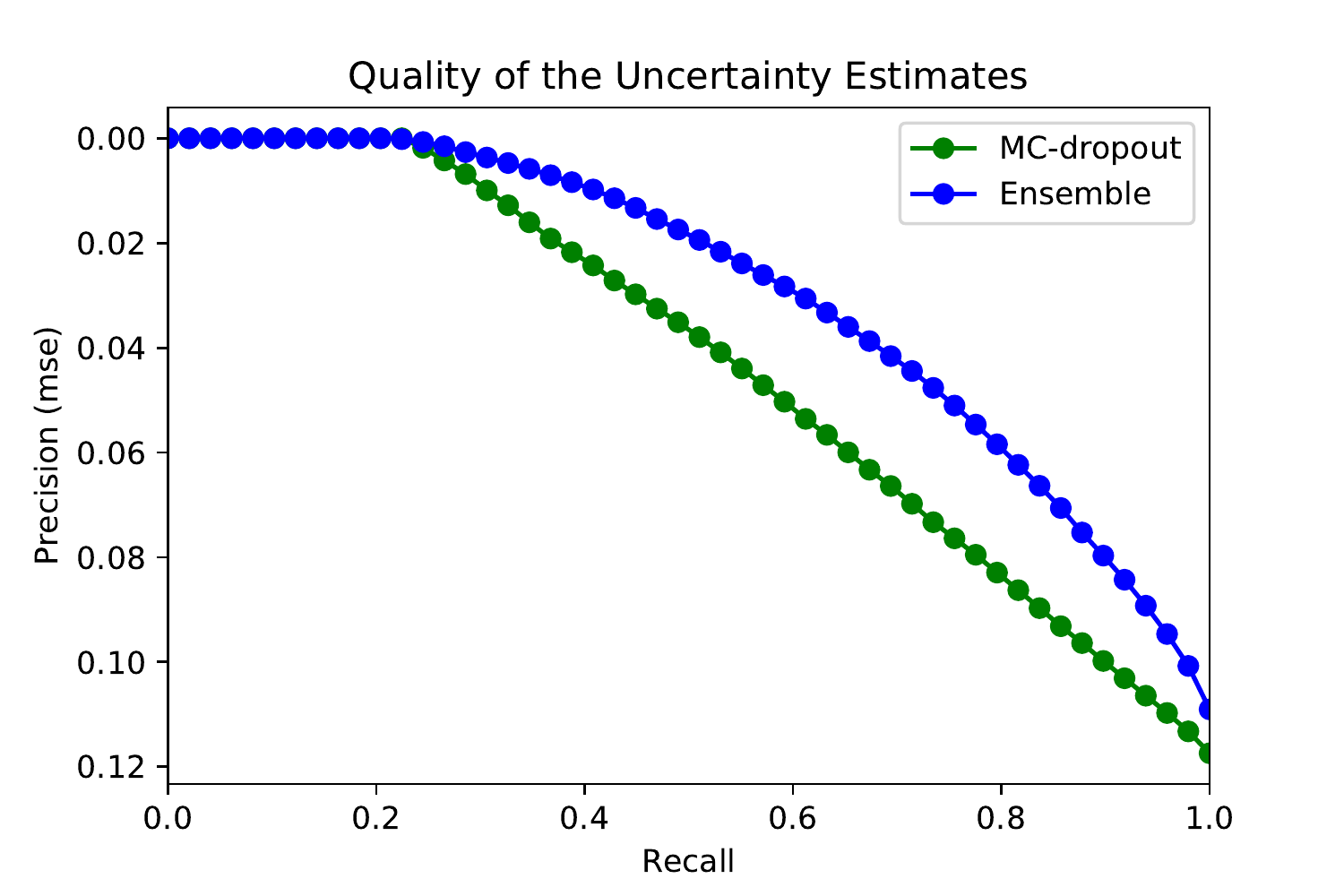}
    \caption{Calibration of MC-dropout and explicit model ensemble. Uncertainty was estimated from an ensemble of 5 models, respectively 5 forward passes.}
    \label{fig:uncertainty_calibration}
\end{figure}

Uncertainty estimates are only useful if they are reasonably well calibrated with respect to the validation error we might expect.
In Figure \ref{fig:uncertainty_calibration}, we constructed a precision-recall curve based on the predicted uncertainty and the residual values from validation data.
The curve shows how the average MSE reduces by removing samples with predicted uncertainty above percentile thresholds. Aligned with the related works, we can see that the estimates of an ensemble of 5 models yield much better calibrated estimates for the uncertainty than a MC-dropout.

%% file: algorithm_1.tex
\begin{algorithm}
  \caption{Large-scale Active Learning Implementation}
  \textbf{Require:} ensemble of $T$ trained models $F_t$, $M$ machines, annotation budget $B$, Unlabelled dataset $\mathcal{U}$ \\
  \textbf{Output:} Set \add{$\mathcal{A} = \{x^*\}$} \remove{$\mathcal{L}_\text{add}$} with $B$ samples to be labelled by human experts.
  \begin{algorithmic}[1]
  \State Split $\mathcal{U}$ in $Q$ regions.
    \For{\textbf{each} region $q \in Q$} \Comment{In parallel on $M$ machines}
        \For {\textbf{each} sample $x \in q$}
            \State Obtain predictions $\hat{y} = \sum_t F_t(x) / T$
            \State Compute uncertainty $s= \sum_t (\hat{y} - F_t(x))^2 $
            \State Obtain pixel count in $n_x$
            \State Obtain deep embedding statistics
            \Statex \hspace{\algorithmicindent} \hspace{\algorithmicindent} $v = \sum_t F_t^z(x) $ and $w  = \sum_t F_t^z(x)$
        \EndFor
        \State Compute region statistics
        \Statex \hspace{\algorithmicindent} $s_q = \sum_{x \in q} s, n_q = \sum_{x \in q} n_x , v_q = \sum_{x \in q} v, w_q = \sum_{x \in q} w$
    \EndFor
     \State From each region, retrieve $s_q$, $n_q$, $v_q$ and $w_q$
     \State Compute total pixel count and global mean
     \Statex $N = \sum n_q, {\mu} = \sum v_q / N$
    \For{\textbf{each} region $q \in Q$}
        \State Compute total distance to the global mean 
        \Statex \hspace{\algorithmicindent} $d^z(x_q,\mu)= w_q - 2v_q\mu + N\mu^2$
    \EndFor
    \State For each region $g(x_h) = \dfrac{s_q}{\sum_q s_q}  + \dfrac{d^z(x_q,\mu)}{\sum_q d^z(x_q,\mu)}$ 
        \State Construct set $C$ with the $q\cdot B$ regions with highest $g$ score
        \State Cluster set $C$ into $B$ clusters with weighted $k$-means,
        \Statex \hspace{\algorithmicindent} where each region has weight $1/(q\cdot B\cdot g(x_i))$
        \State Construct set \add{$\mathcal{A} = \{x^*\}$} \remove{$\mathcal{L}_\text{add}$} by obtaining the samples $x^*$ closest to 
        \Statex \hspace{\algorithmicindent} each of the $B$ centroids.
  \end{algorithmic}
    \label{alg:core-setconstruction}
\end{algorithm}

%% file: tables/architecture.tex
\begin{tabular}{ll}
\toprule

\multicolumn{2}{l}{Network $F(x) \mapsto \hat{y}$} \\
& Input $\mapsto$ BlockIn $\mapsto$ ResNetBlocks(15)  $\mapsto$  BlockOut $\mapsto (\text{output}_{sem},\text{output}_{reg} )$ \\ 
\midrule
\multicolumn{2}{l}{BlockIn} \\
& Input $\mapsto$ Block(64,64,128)  $\mapsto$ Conv2D(128,3) $\mapsto$  Conv3D(256,3)  \\
\midrule
\multicolumn{2}{l}{ ResNetBlocks(N)} \\
& Input $\mapsto$  ReLU(Input + Block(64,64,256)) $\mapsto$ output  (N times)\\
\midrule
\multicolumn{2}{l}{ Block($n_1,n_2,n_3$)} \\
& Input $\mapsto$  Conv2D($n_1$,3) $\mapsto$  Conv2D($n_2$,3) $\mapsto$  Conv2D($n_3$,3) $\mapsto$ output \\
\midrule
\multicolumn{2}{l}{ BlockOut} \\
& Input $\mapsto$  Conv2D(2,3)  $\mapsto \text{output}_{cl}$ \\
& Input $\mapsto$  Conv2D(1,3)  $\mapsto \text{output}_{reg}$ \\
\bottomrule
\end{tabular}